\documentclass{ieeetran}
\renewcommand{\baselinestretch}{0.97}
\usepackage[american]{babel}
\renewcommand{\baselinestretch}{0.97}
\usepackage{natbib} % has a nice set of citation styles and commands
\usepackage{mathtools} % amsmath with fixes and additions
\usepackage{amssymb,amsthm}
\usepackage{booktabs} % commands to create good-looking tables
\usepackage{tikz} % nice language for creating drawings and diagrams

\newcommand\scalemath[2]{\scalebox{#1}{\mbox{\ensuremath{\displaystyle #2}}}}
\usepackage{xcolor}
\usepackage{hyperref}       % hyperlinks
\usepackage{url}            % simple URL typesetting
\usepackage{booktabs}       % professional-quality tables
\usepackage{amsfonts}       % blackboard math symbols
\usepackage{nicefrac}       % compact symbols for 1/2, etc.
\usepackage{microtype}      % microtypography
\usepackage{xcolor}         % colors
\usepackage{wrapfig}
\usepackage{dsfont}
\usepackage{amsmath,amssymb,amsfonts}
\usepackage{algorithm}
\usepackage{algorithmicx}
\usepackage{algpseudocode}
\usepackage{caption}
\usepackage{amsmath, amssymb, bbm, xspace}
\usepackage{epsfig}
\usepackage{longtable}
\usepackage{color}
\usepackage{mathrsfs}
\usepackage{comment}

\usepackage{courier}
\newtheorem{theorem}{Theorem}[section]
\newtheorem{assumption}{Assumption}[section]

\newtheorem{lemma}{Lemma}[section]

\newtheorem{definition}{Definition}[section]

\def\bkE{{\rm I\kern-.17em E}}
\def\bk1{{\rm 1\kern-.17em l}}
\def\bkD{{\rm I\kern-.17em D}}
\def\bkR{{\rm I\kern-.17em R}}
\def\bkP{{\rm I\kern-.17em P}}

\def\bkZ{{\bf{Z}}}

\def\bkE{{\rm I\kern-.17em E}}
\def\bk1{{\rm 1\kern-.17em l}}
\def\bkD{{\rm I\kern-.17em D}}
\def\bkR{{\rm I\kern-.17em R}}
\def\bkP{{\rm I\kern-.17em P}}

\makeatletter
\newcommand{\pushright}[1]{\ifmeasuring@#1\else\omit\hfill$\displaystyle#1$\fi\ignorespaces}
\newcommand{\pushleft}[1]{\ifmeasuring@#1\else\omit$\displaystyle#1$\hfill\fi\ignorespaces}
\makeatother

\def\bkZ{{\bf{Z}}}
\def\b12{(\beta_1,\beta_2)}

\newcounter{example}
\renewcommand{\theexample}{\thesection.\arabic{example}}

\newcounter{remark}
\renewcommand{\theremark}{\thesection.\arabic{remark}}

\def\Hscr{\mathscr{H}}

\def\Xscr{\mathcal{X}}

\def\Ebb{\mathbb{E}}
\newlength{\noteWidth}
\long\def\notes#1{\ifinner
{\tiny #1}
\else
\marginpar{\parbox[t]{\noteWidth}{\raggedright\tiny #1}}
\fi\typeout{#1}}

 \def\notes#1{\typeout{read notes: #1}} %uncomment for final version

\def\Hcal{\mathcal{H}}

\def\Ebb{\mathbb{E}}

\def\Ibb{{\mathbb{I}}}

\def\ast{\buildrel{t}\over\rightarrow}

\def\exp{\mathop{\hbox{\rm exp}}}

\def\spose#1{\hbox to 0pt{#1\hss}}

\def\text #1{\hbox{\quad#1\quad}}

\def\Escr{\mathcal{E}}

\def\nthinsp{\mskip -2   mu}

\def\superstar{^{\raise 0.5pt\hbox{$\nthinsp *$}}}
\def\SUPERSTAR{^{\raise 0.5pt\hbox{$*$}}}
\def\lamstarT {\lambda^{\raise 0.5pt\hbox{$\nthinsp *$}T}}

\def\Ascr{{\cal A}}

\def\Nscr{{\cal N}}

\def\Xscr{{\cal X}}

\def\chat{\skew3\widehat c}

\def\non{\nonumber}

\let\forallnew\forall
\renewcommand{\forall}{\forallnew\ }
\let\forall\forallnew

		\def\bkE{{\rm I\kern-.17em E}}
		\def\bk1{{\rm 1\kern-.17em l}}
		\def\bkD{{\rm I\kern-.17em D}}
		\def\bkR{{\rm I\kern-.17em R}}
		\def\bkP{{\rm I\kern-.17em P}}
		\def\bkY{{\bf \kern-.17em Y}}
		\def\bkZ{{\bf \kern-.17em Z}}
		\def\bkC{{\bf  \kern-.17em C}}
{\begin{list}{}%
         {\setlength{\leftmargin}{#1}}%
         \item[]%
}
{\end{list}}

		\def\bsp{\begin{split}}
		\def\beq{\begin{eqnarray}}
		\def\bal{\begin{align*}}
		\def\bc{\begin{center}}
		\def\be{\begin{enumerate}}
		\def\bi{\begin{itemize}}
		\def\bs{\begin{small}}
		\def\bS{\begin{slide}}
		\def\ec{\end{center}}
		\def\ee{\end{enumerate}}
		\def\ei{\end{itemize}}
		\def\es{\end{small}}
		\def\eS{\end{slide}}
		\def\eeq{\end{eqnarray}}
		\def\eal{\end{align*}}
		\def\esp{\end{split}}
		\def\qed{ \vrule height7.5pt width7.5pt depth0pt}  %width4.17pt depth0pt} 

	\def\cp2problem#1#2#3#4{\fbox
		 {\begin{tabular*}{0.9\textwidth}
			{@{}l@{\extracolsep{\fill}}l@{\extracolsep{6pt}}l@{\extracolsep{\fill}}c@{}}
				#1 & & $#4 $ 
			\end{tabular*}}}

		\def\bkE{{\rm I\kern-.17em E}}
		\def\bk1{{\rm 1\kern-.17em l}}
		\def\bkD{{\rm I\kern-.17em D}}
		\def\bkR{{\rm I\kern-.17em R}}
		\def\bkP{{\rm I\kern-.17em P}}
		
		\def\bkZ{{\bf{Z}}}

\newcommand {\beeq}[1]{\begin{equation}\label{#1}}
\newcommand {\eeeq}{\end{equation}}
\newcommand {\bea}{\begin{eqnarray}}
\newcommand {\eea}{\end{eqnarray}}

\def\texitem#1{\par\smallskip\noindent\hangindent 25pt
               \hbox to 25pt {\hss #1 ~}\ignorespaces}

\def\bsp{\begin{split}}
		\def\beq{\begin{eqnarray}}
		\def\bal{\begin{align*}}
		\def\bc{\begin{center}}
		\def\be{\begin{enumerate}}
		\def\bi{\begin{itemize}}
		\def\bs{\begin{small}}
		\def\bS{\begin{slide}}
		\def\ec{\end{center}}
		\def\ee{\end{enumerate}}
		\def\ei{\end{itemize}}
		\def\es{\end{small}}
		\def\eS{\end{slide}}
		\def\eeq{\end{eqnarray}}
		\def\eal{\end{align*}}
		\def\esp{\end{split}}
		\def\qed{ \vrule height7.5pt width7.5pt depth0pt}  %width4.17pt depth0pt} 

\usepackage{amsmath, amssymb, xspace}
\usepackage{epsfig}
\usepackage{longtable}
\usepackage{color}
\usepackage{mathrsfs}
\usepackage{subfig}

\def\Nscr{{\cal N}}

\def\Ascr{{\cal A}}
\usepackage{graphicx}
\usepackage{textcomp}
\usepackage{xcolor}
\usepackage{comment}
\usepackage{caption}
\renewcommand\scalemath[2]{\scalebox{#1}{\mbox{\ensuremath{\displaystyle #2}}}}

\newcounter{protocol}
\makeatletter
\newenvironment{protocol}[1][htb]{%
  \let\c@algorithm\c@protocol
  \renewcommand{\ALG@name}{Protocol}% Update algorithm name
  \begin{algorithm}[#1]%
  }{\end{algorithm}
}
\makeatother
\def \Acal {\mathcal{A}}
\def \Ccal {\mathcal{C}}
\def \Rcal {\mathcal{R}}
\def \Hscr {\mathcal{H}}
\def \gammastar {\gamma^{*}}
\def \thetastar {\theta^{*}}
\def \chat {\hat{c}}
\def \Ebb {\mathbb{E}}
\def \Nscr {\mathcal{N}}
\def \Fcal {\mathcal{F}}
\def \Ibb{\mathbb{I}}
\def \indicator{\boldsymbol{1}}

\def \thetastart{\theta^{*\top}}

\newcommand {\Tone}{{\mathrm{Term}_1}}
\newcommand {\Ttwo}{{\mathrm{Term}}_2}
\newcommand {\Tthree}{{\mathrm{Term}}_3}

\def \thetastart{\theta^{*\top}}
\def \gammastart{\gamma^{*\top}}
\bibpunct[, ]{(}{)}{;}{a}{,}{,}
\title{Thompson Sampling for Stochastic Bandits with Noisy Contexts: An Information-Theoretic Regret Analysis}

\author{Sharu Theresa Jose and Shana Moothedath% <-this % stops a space
\thanks{S. T. Jose is with Department of Computer Science,
        University of Birmingham, Birmingham -B15 2TT, UK
        {\tt\small s.t.jose@bham.ac.uk}}%
\thanks{S. Moothedath is with the Department of Electrical Engineering, Iowa State University,
        Ames, IA 50011, USA
        {\tt\small mshana@iastate.edu}}%
}

\begin{document}
\maketitle
\begin{abstract}
%{\em We investigate a stochastic contextual linear bandit problem, where the environment generates context vectors independently based on a known distribution. The agent, however, does not directly observe the true context; instead, it receives a noisy, corrupted version of the context through a noise channel with an unknown noise parameter. The agent then selects an action and receives a reward.} 
 We study stochastic linear contextual bandits (CB) where the agent observes a \textit{noisy} version of the true context through a noise channel with unknown channel parameter.
Our objective is to design an action policy that can ``approximate" that of a Bayesian oracle that  has access to the reward model and the noise channel parameter.
%The oracle uses the predictive distribution of the true context given the observed noisy context and the mean channel noise to select an action. 
We introduce a modified Thompson sampling  algorithm and analyze its Bayesian cumulative regret with respect to the oracle action policy via information-theoretic tools.
%In a Bayesian framework, we introduce a Thompson sampling algorithm for Gaussian bandits with Gaussian context noise. 
For $d$-dimensional Gaussian bandits with Gaussian context noise, our information-theoretic analysis shows that the Bayesian cumulative regret  
scales as $O(d\sqrt{T})$, where $T$ is the horizon. We also consider the problem setting where the agent observes the true context with some delay after receiving the reward, and show that delayed true contexts lead to lower regret. Finally, we empirically demonstrate the performance of the proposed
algorithms against baselines.
%We consider as baseline the action policy of an oracle that has access to the reward model, as well as the predictive distribution of true context from the observed noisy context. Considering a Bayesian framework, we propose a Thompson sampling-based algorithm for Gaussian bandits with Gaussian context noise. We leverage information-theoretic tools to show that the Bayesian regret of the proposed algorithm with respect to the oracle action policy scales as $O(d\sqrt{T})$ where $d$ is the dimension of the context vector. We also extend the problem to a delayed context setting when the agent observes the true context {\cblue with some delay, specifically} after it receives the reward. The availability of delayed true contexts is shown to ensure a lower Bayesian regret. Finally, we demonstrate the performance of the proposed algorithms on a  synthetic dataset.
\end{abstract}

\section{Introduction}
Decision-making in the face of uncertainty is a widespread challenge found across various domains such as control and robotics \citep{srivastava2014surveillance}, clinical trials \citep{aziz2021multi}, communications \citep{anandkumar2011distributed}, and ecology \citep{srivastava2013optimal}.  
To tackle this challenge, learning algorithms have been developed to uncover effective policies for optimal decision-making. One notable framework for addressing this is contextual bandits (CBs), which captures the essence of sequential decision-making by incorporating side information, termed {\it context} \citep{bubeck2012regret}. 
In the standard CB model, an agent  interacts with the environment over numerous rounds. 
%{\color{blue}In each round, the environment presents a context to the learner, who subsequently chooses an action based on the observed context. After this selection, the learner receives a reward associated with the chosen action, with the overarching aim of maximizing cumulative rewards.} 
 In each round, the environment presents a context to the agent based on which the agent chooses an action and receives a reward from the environment. The goal of the agent is to design a policy for action selection that can maximize the cumulative mean reward accrued over a $T$-length horizon.

While most prior research on CBs has primarily focused on models with known exact contexts \citep{auer2002finite, chu2011contextual,agrawal2013thompson}, in many real-world applications, the contexts are noisy, e.g. imprecise measurement of patient conditions in clinical trials, weather or stock market predictions. In such scenarios when the exact contexts are unknown, the agent must utilize the observed noisy contexts to estimate the mean reward associated with the true context. However, this results in a biased estimate that renders the application of standard CB algorithms unsuitable. Consequently, recent efforts have been made to develop CB algorithms tailored to noisy context settings.
%and instead, we have access only to a context distribution \citep{kirschner2019stochastic} or to noisy contexts \citep{kim2023contextual, guo2023online, yang2020multi,lamprier2018profile}. 
%Estimating the mean reward using the observed noisy contexts instead of the true contexts results in a biased estimate that renders application of standard CB algorithms unsuitable.

{\emph{Related Works}:}  \citep{lamprier2018profile} considers a setting where there is a bounded zero-mean noise in the \textit{feature vector} (denoted by $\phi(a,c)$, where $a$ is the action and $c$ is the context) rather than in the context vector, and the agent observes only noisy features. For this setting, they develop an upper confidence bound (UCB) algorithm.  \citep{kirschner2019stochastic} models the uncertainty regarding the true contexts by a \textit{context distribution} that is known to the agent, while the agent never observes the true context; and develops a UCB algorithm. A similar setting has also been considered in \citep{yang2020multi}. Different from these works, \citep{kim2023contextual} considers the setting where the true feature vectors are sampled from an unknown feature distribution at each time, but the agent observes only a noisy feature vector. Assuming Gaussian feature noise with unknown mean and covariance, they develop an OFUL algorithm. A variant of this setting has been studied in \citep{guo2023online}. 
%In particular, frequentist, upper confidence bound (UCB) algorithms have been developed in \citep{kirschner2019stochastic} for when only context distribution is known \citep{kirschner2019stochastic}, and in \citep{kim2023contextual, guo2023online, lamprier2018profile} for when only noisy context is observed.

\begin{table*}[t]
  \centering
  \begin{tabular}{|c|c|c|c|c|}
  \hline
    Reference & Setting & Algorithm & Regret & Bound \\
    \hline
    \citep{auer2002finite} / \citep{chu2011contextual} & Linear CB & LinRel/ Lin-UCB & Frequentist & $\tilde{O}(\sqrt{dT})$ \\ \hline
    \citep{agrawal2013thompson} & Linear CB & TS & Frequentist & $O(d\sqrt{T} \log^{3/2} T)$ \\ \hline 
    \citep{russo2014learning} & Linear CB & TS & Bayesian & $O(d\sqrt{T} \log T)$\\ \hline 
    \citep{lamprier2018profile}/ \citep{kirschner2019stochastic} & Noisy CB & SampLinUCB/UCB & Frequentist & $\tilde{O}(d\sqrt{T})$ \\ \hline
    \citep{kim2023contextual} & Noisy CB & OFUL & Frequentist & $\tilde{O}(d\sqrt{T})$ \\ \hline
    \textbf{Our work} & \textbf{Noisy CB} & \textbf{TS} &\textbf{ Bayesian} & \textbf{$O(d\sqrt{T})$} \\ \hline 
  \end{tabular}
  \caption{ Comparison of the regret bounds of our proposed TS algorithm for noisy CB with state-of-the art algorithms.}
  \label{tab:1}
\end{table*}
%\end{footnotesize}

\emph{Motivation and Problem Setting:} In this work, inspired by \citep{kim2023contextual}, we consider the following noisy CB setting. In each round, the environment samples a true context vector $c_t$ from a \textit{context distribution} that is \textit{known} to the agent. The agent, however, does not observe the true context, but observes a noisy context $\chat_t$ obtained as the output of a noise channel $P(\chat_t|c_t,\gammastar)$ parameterized by $\gammastar$. The agent is aware of the noise present but does not know the channel parameter $\gammastar$. Following \citep{kim2023contextual},  we consider Gaussian noise channels for our regret analysis.

Based on the observed noisy contexts, the agent chooses an action $a_t$ and observes a reward  $r_t$ corresponding to the true context. We consider a linear bandit whose mean reward $\phi(a_t,c_t)^{\top}\thetastar$ is determined by an unknown reward parameter $\thetastar$. The goal of the agent is to design an action policy that minimizes the \textit{Bayesian cumulative regret} with respect to the action policy of a Bayesian oracle. The oracle has access to the
reward model and the channel parameter $\gammastar$,  and uses the predictive distribution
of the true context given the observed noisy context
to select an action. 
%The assumption of Gaussian noise is in line with \cite{kim2023contextual}.

Our setting differs from \citep{kim2023contextual} in that we assume noisy contexts rather than noisy feature vectors and that the agent knows the context distribution.  The noise model, incorporating noise in the feature vector, allows  \citep{kim2023contextual} to transform the original problem into a different CB problem that estimates a modified reward parameter. 
%An OFUL-based algorithm with frequentist regret guarantees is proposed. 
 Such a transformation however is not straightforward in our setting with noisy contexts, where we wish to analyze the Bayesian regret. Additionally, we propose a de-noising approach to estimate the predictive distribution of the true context from given noisy contexts,  offering potential benefits for future analyses.

The assumption of known context distribution follows from \citep{kirschner2019stochastic}. This can be motivated by considering the example of an online recommendation engine that pre-process the user account registration information or contexts (e.g., age, gender, device, location, item preferences) to group them into different clusters \citep{roy2022systematic}. The engine can then infer the `empirical' distribution of users within each cluster to define a context distribution over true contextual information. A noisy contextual information scenario occurs when a guest with different preferences logs into a user’s account.

 \emph{Challenges and Novelty:} Different from existing works that developed UCB-based algorithms, we propose a fully Bayesian TS algorithm that approximates the Bayesian oracle policy. The proposed algorithm differs from the standard contextual TS \citep{agrawal2013thompson} in the following aspects. Firstly, since the true context vectors are not accessible at each round and the channel parameter $\gammastar$ is unknown, the agent uses its knowledge of the context distribution and the past observed noisy contexts to infer a \textit{predictive posterior} distribution of the true context from the current observed noisy context.  The inferred predictive distribution is then used to choose the action. This \textit{de-noising} step enables our algorithm to `approximate' the oracle action policy that uses knowledge of the channel parameter $\gammastar$ to implement {\it exact de-noising}.
Secondly, the reward $r_t$ received by the agent corresponds to the unobserved true context $c_t$. Hence, the agent cannot accurately evaluate the  posterior distribution of $\thetastar$ and sample from it as is done in standard contextual TS. Instead, our algorithm proposes to use a sampling distribution that `approximates' the posterior. 

Moreover, different from existing works that focus on frequentist regret analysis, we derive novel \textit{information-theoretic} bounds on the \textit{Bayesian cumulative regret} of our algorithm. For Gaussian bandits, our information-theoretic regret bounds scale as $O(d\sqrt{T})$ where $d, T$ denotes the context vector's dimension and time horizon, respectively. Furthermore, our Bayesian regret analysis shows that the \textit{posterior mismatch}, resulting due to replacing the true posterior distribution with a sampling distribution, results in an approximation error that is captured via the Kullback-Leibler (KL) divergence between the distributions. 
We also
extend our algorithm to a setting where the agent
observes the true context after a delay following the reward reception \citep{kirschner2019stochastic}, and show that delayed true contexts
result in reduced Bayesian regret.
Table~\ref{tab:1} compares our regret bound with that of the state-of-the-art algorithms in the noiseless and noisy CB settings.

\section{Problem Setting}\label{sec:problemformulation}
In this section, we present the stochastic linear CB problem studied in this paper. Let $\Acal$ denote the action set with $K$  actions and $\Ccal$ denote the (possibly infinite) set of $d$-dimensional context vectors. At iteration $t\in \mathbb{N}$, the environment randomly draws a context vector $c_t \in \Ccal$ according to a {\it context distribution} $P(c)$ defined over the space $\Ccal$ of context vectors. The context distribution $P(c)$ is known to the agent. The agent, however, does not observe the true context $c_t$ drawn by the environment. Instead, it observes a noisy version $\chat_t$ of the true context, obtained as the output of a noisy, stochastic channel  $P(\chat_t|c_t,\gammastar)$ with the true context $c_t$ as the input. The noise channel $P(\chat_t|c_t,\gammastar)$ is parameterized  by the {\it noise channel parameter} $\gammastar$ that is {\it unknown} to the agent. 

Having observed the noisy context $\chat_t$ at iteration $t$, the agent chooses an action $a_t \in \Acal$ according to an {\it action policy} $\pi_t(\cdot|\chat_t)$. The action policy may be stochastic describing a probability distribution over the set $\Acal$ of actions.
Corresponding to the chosen action $a_t$, the agent receives a  reward from the environment given by \begin{align}r_t = f(\thetastar,a_t,c_t)+\xi_t, \label{eq:rewardfunction}\end{align} where $f(\thetastar,a_t,c_t)=\phi(a_t,c_t)^\top\thetastar$  is the linear {\it mean-reward function} and $\xi_t$ is a zero-mean reward noise variable. The mean reward function $f(\thetastar,a_t,c_t)$ is defined via the {\it feature map} $\phi: \Acal \times \Ccal \rightarrow \mathbb{R}^{m}$, that maps the action and true context to an $m$-dimensional feature vector, and via the reward parameter $\thetastar \in \mathbb{R}^{m}$ that is {\it unknown} to the agent.

We call the noisy CB problem described above {\it CBs with unobserved true context} (see Setting~\ref{problem:2}) since the agent does not observe the true context $c_t$ and the selection of action is based solely on the observed noisy context. Accordingly, at the end of iteration $t$, the agent has accrued the history $\Hscr_{t,r,a,\chat}=\{r_{\tau},a_{\tau},\chat_{\tau}\}_{\tau=1}^t$ of observed reward-action-noisy context tuples. The action policy $\pi_{t+1}(\cdot|\chat_{t+1})$ at $(t+1)^{\rm th}$ iteration may depend on the history $\Hscr_{t,r,a,\chat}$. 

\begin{protocol}[h!]
\caption{\textbf{Setting 1}: CBs with unobserved true contexts}\label{problem:2}
\begin{algorithmic}[1]
\For{$t=1,\hdots,T$}
\State Environment samples $c_t \sim P(c)$.
\State Agent observes noisy context $\chat_t \sim P(\chat_t|c_t,\gammastar)$.
\State Agent chooses an action $a_t \sim \pi_t(\cdot|\chat_t)$.
\State Agent receives reward $r_t$ according to \eqref{eq:rewardfunction}.
\EndFor
\end{algorithmic}
\end{protocol}

We also consider a variant of the above problem setting where the agent has access to a \textit{delayed} observation of the true context $c_t$ as studied in \citep{kirschner2019stochastic}. We call this setting \textit{CBs with delayed true context}. In this setting, at iteration $t$, the agent observes the true context $c_t$ {\it after} it receives reward $r_t$ corresponding to the action $a_t \sim \pi_t(\cdot|\chat_t)$ chosen based on the observed noisy context $\chat_t$. It is important to note that the agent has no access to the true context at the time of decision-making. 
%Due to the delayed observation of true context, the agent's choice of action is determined only by the observed noisy context. 
%while the true context $c_t$ is not revealed to the agent during the time of selecting the action, the agent observes it \textit{after} it receives the reward $r_t$ corresponding to the chosen action $a_t$.
Thus, at the end of iteration $t$, the agent has collected the history $\Hscr_{t,r,a,c,\chat}=\{r_{\tau},a_{\tau},c_{\tau},\chat_{\tau}\}_{\tau=1}^t$ of observed reward-action-context-noisy context tuples.

In both of the problem settings described above, the agent's objective is to devise an action policy that minimizes the {\em Bayesian cumulative regret} with respect to a baseline action policy. We define Bayesian cumulative regret next.

 \subsection{Bayesian Cumulative Regret}
 The cumulative regret of an action policy $\pi_t(\cdot|\chat_t)$ quantifies how different the mean reward accumulated over $T$ iterations is from that accrued by a  baseline action policy $\pi^{*}_t(\cdot|\chat_t)$. In this work, we consider as baseline the action policy of an {\it oracle} that has access to the channel noise parameter $\gammastar$, reward parameter $\thetastar$, the context distribution $P(c)$ and the noise channel likelihood $P(c_t|\chat_t,\gammastar).$ Accordingly, at each iteration $t$, the oracle can infer the {\it exact predictive distribution} $P(c_t|\chat_t,\gammastar)$ of the true context from the observed noisy context $\chat_t$ via Baye's rule as \begin{align}P(c_t|\chat_t,\gammastar)=\frac{P(c_t,\chat_t|\gammastar)}{P(\chat_t|\gammastar)}\label{eq:c_chatgammastar}.\end{align} Here, $P(c_t,\chat_t|\gammastar)=P(c_t)P(\chat_t|c_t,\gammastar)$ is the joint distribution of the true and noisy contexts given the noise channel parameter $\gammastar$,  and $P(\chat_t|\gammastar)$ is the distribution obtained by marginalizing $P(c_t,\chat_t|\gammastar)$ over the true contexts,
 {\emph{i.e.}, \begin{align}
 P(\chat_t|\gammastar) =\Ebb_{P(c_t)}[P(\chat_t|c_t,\gammastar)] \label{eq:chat_gammastar},
 \end{align}where $\Ebb_{\bullet}[\cdot ]$ denotes expectation with respect to $`\bullet$'.
 The oracle action policy then adopts an action 
 \begin{align}
     a^*_t &= \arg \max_{a \in \Acal} \Ebb_{P(c_t|\chat_t,\gammastar)}[\phi(a,c_t)^\top\thetastar] \nonumber \\
     & =\arg \max_{a \in \Acal} \psi(a,\chat_t|\gammastar)^\top\thetastar, \label{eq:oracle_action}
 \end{align} at iteration $t$, where  $\psi(a,\chat|\gammastar):=\Ebb_{P(c|\chat,\gammastar)}[\phi(a,c)]$. Note that as in \citep{kim2023contextual,park2021analysis}, we do not choose the stronger oracle action policy of $\arg \max_{a \in \Ascr} \phi(a,c_t)^\top\thetastar$, that requires access to the true context $c_t$, as it is generally not achievable by an agent that observes only noisy context $\chat_t$  and has  no access to  $\gammastar$. 

 For fixed parameters $\thetastar$ and $\gammastar$, we define the cumulative regret of the action policy $\pi_t(\cdot|\chat_t)$ as \begin{align}
\hspace{-1 mm}\scalemath{0.95}{\hspace{-2 mm} \mathcal{R}^T (\pi|\thetastar,\gammastar)\hspace{-1.0 mm}=\hspace{-1.0 mm}\mathbb{E}\Bigl[\sum_{t=1}^T \hspace{-1 mm}\phi(a^{*}_t,c_t)^\top \thetastar \hspace{-1.0 mm}- \hspace{-1.0 mm}\phi(a_t,c_t)^\top\thetastar \bigl| \thetastar,\gammastar\Bigr]},\label{eq:cum_regret}
\end{align} the expected difference in mean rewards of the oracle decision policy and the agent's decision policy over $T$ iterations. The expectation is taken over the randomness in the selection of actions $a^*_t$ and $a_t$, as well as true context $c_t$.
Equivalently, the cumulative regret of \eqref{eq:cum_regret} can be written as
\begin{align}
 &\mathcal{R}^T(\pi|\thetastar,\gammastar) \nonumber \\
 &=\hspace{-1 mm}\sum_{t=1}^T\mathbb{E}\bigl[ \Ebb \bigl[\phi(a^{*}_t,c_t)^\top \thetastar-\phi(a_t,c_t)^\top\thetastar \bigl | \chat_t,a_t \bigr] \bigl | \thetastar,\gammastar\bigr]\nonumber \\
 & =\hspace{-1 mm}\sum_{t=1}^T\mathbb{E}\bigl[  \psi(a^{*}_t,\chat_t|\gammastar)^\top \thetastar \hspace{-0.7 mm}-\hspace{-0.7 mm}\psi(a_t,\chat_t|\gammastar)^\top\thetastar   \bigl| \thetastar,\gammastar\bigr].\label{eq:cum_regret_redefined}
\end{align} 
Our focus in this work is on a {\it Bayesian framework} where we assume that the reward parameter $\thetastar \in \Theta$ and channel noise parameter $\gammastar \in \Gamma$  are independently sampled by the environment from prior distributions $P(\thetastar)$, defined on the set $\Theta$ of reward parameters, and $P(\gammastar)$, defined on the set $\Gamma$ of channel noise parameters, respectively. The agent has knowledge of the prior distributions, the reward likelihood in \eqref{eq:rewardfunction} and the noise channel likelihood $P(\chat_t|c_t,\gammastar)$, although it does not observe the sampled $\gammastar$ and $\thetastar$. Using the above prior distributions, we define {\it Bayesian cumulative regret} of the action policy $\pi_t(\cdot|\chat_t)$ as
\begin{align}
    \Rcal^T(\pi)=\Ebb[\Rcal^T(\pi|\thetastar,\gammastar)], \label{eq:Bayesian_cumregret}
\end{align} where the expectation is taken with respect to the priors $P(\thetastar)$ and $P(\gammastar)$.
  
In the next sections, we present our novel TS algorithms to minimize the Bayesian cumulative regret for the two problem settings considered in this paper. 
%To this end, we focus on linear-Gaussian stochastic CBs with Gaussian context noise as described next. 
\section{Modified TS for CB with Unobserved True Contexts}
In this section, we consider Setting~\ref{problem:2} where the agent only observes the noisy context $\chat_t$ at each iteration $t$. Our proposed modified TS Algorithm is given in Algorithm~\ref{alg:2}.

\begin{algorithm}[h!]
\caption{\textbf{Algorithm 1}: TS with unobserved true contexts ($\pi^{\rm{TS}}$)}\label{alg:2}
\begin{algorithmic}[1]
\For{$t=1,\hdots,T$}
\State The environment selects a true context $c_t$.
\State Agent observes noisy context $\chat_t$.
\State Agent evaluates the predictive posterior distribution $P(c_t|\chat_t,\Hscr_{t-1,\chat})$ as in \eqref{eq:predictiveposterior_1}.
\State Agent samples $\theta_t \sim \bar{P}(\thetastar|\Hscr_{t-1,r,a,\chat})$
\State Agent chooses action $a_t$ as in \eqref{eq:actionpolicy_nocontexts}.
\State Agent observes reward $r_t$ as in \eqref{eq:rewardfunction}.
%\State 7. Update the history $\Hscr_{t,r,a,c,\chat}=\Hscr_{t-1,r,a,c,\chat} \cup (r_t,a_t,c_t,\chat_t)$
\EndFor
\end{algorithmic}
\end{algorithm}

The proposed algorithm implements two steps in each iteration $t \in \mathbb{N}$. In the first step, called the {\it de-noising} step, the agent uses the current observed noisy context $\chat_t$ and the history $\Hscr_{t-1,\chat}=\{\chat_{\tau}\}_{\tau=1}^{t-1}$ of past observed noisy contexts to obtain a \textit{predictive posterior distribution} $P(c_t|\chat_t,\Hscr_{t-1,\chat})$ of the true context $c_t$. This is a two-step process, where firstly the agent uses the history $\Hscr_{t-1,\chat}$ of past observed noisy contexts to compute the posterior distribution of $\gammastar$ as
$
    P(\gammastar|\Hscr_{t-1,\chat}) \propto P(\gammastar) \prod_{\tau=1}^{t-1}P(\chat_{\tau}|\gammastar), $ where the conditional distribution $P(\chat_t|\gammastar)$ is evaluated as in \eqref{eq:chat_gammastar}. Note that to evaluate the posterior, the agent uses its knowledge of the context distribution $P(c)$,  the prior $P(\gammastar)$ and the noise channel likelihood $P(\chat_t|c_t,\gammastar)$.  Using the derived posterior $P(\gammastar|\Hscr_{t-1,\chat})$, the predictive posterior distribution of the true context is then obtained as
\begin{align}
    P(c_t|\chat_t,\Hscr_{t-1,\chat})=\Ebb_{P(\gammastar|\Hscr_{t-1,\chat})}[P(c_t|\chat_t,\gammastar)], \label{eq:predictiveposterior_1}
\end{align} where $P(c_t|\chat_t,\gammastar)$ is defined as in \eqref{eq:c_chatgammastar}.

The second step of the algorithm implements a \textit{modified} Thompson sampling. Note that since the agent does not have access to the true contexts, it cannot evaluate the posterior distribution with known contexts, 
\begin{align}
\scalemath{0.96}{P(\thetastar|\Hscr_{t-1,r,a,c}) \propto P(\thetastar) \prod_{\tau=1}^{t-1} P(r_{\tau}|a_{\tau},c_{\tau},\thetastar)}, \label{eq:posterior}
\end{align} as is done in standard contextual TS. Instead, the agent must evaluate the true \textit{posterior distribution} under noisy contexts,
\begin{align}
P_t(\thetastar)&:= P(\thetastar|\Hscr_{t-1,r,a,\chat}) \label{eq:trueposterior} \\& \scalemath{0.96}{\propto P(\thetastar) \prod_{\tau=1}^{t-1} \Ebb_{P(c_{\tau}|\chat_{\tau},\Hscr_{\tau-1,\chat})}[P(r_{\tau}|a_{\tau},c_{\tau},\thetastar)]}, \nonumber
\end{align} where $\Ebb_{P(c_{\tau}|\chat_{\tau},\Hscr_{\tau-1,\chat})}[P(r_{\tau}|a_{\tau},c_{\tau},\thetastar)]$ is the marginal distribution with respect to the predictive posterior in \eqref{eq:predictiveposterior_1}. 
 However, evaluating the marginal distribution is challenging for linear bandits with mean reward $\phi(a_{\tau},c_{\tau})^{\top}\thetastar$, which results in the posterior $P_t(\thetastar)$ to be intractable in general. Consequently, at each iteration $t$, the agent samples $\theta_t \sim \bar{P}(\thetastar|\Hscr_{t-1,r,a,\chat})$ from a distribution $\bar{P}(\thetastar|\Hscr_{t-1,r,a,\chat})$ that `approximates' the true posterior $P_t(\thetastar)$. Specific choice of this sampling distribution depends on the problem setting. Ideally one must choose a distribution that is sufficiently `close' to the true posterior. In the next sub-section, we will explain the choice for Gaussian bandits.

Using the sampled $\theta_t$ and the predictive posterior distribution $P(c_t|\chat_t,\Hscr_{t-1,\chat})$ obtained from the denoising step, the agent then chooses action $a_t$ at iteration $t$ as 
\begin{align}
    a_t &=\arg \max_{a \in \Acal} \psi(a,\chat_t|\Hscr_{\chat})^\top\theta_t,\label{eq:actionpolicy_nocontexts} \hspace{0.2cm} \mbox{where}
\\ \psi(a_t,\chat_t|\Hscr_{\chat})&:=\Ebb_{P(c_t|\chat_t,\Hscr_{t-1,\chat})}[\phi(a_t,c_t)] \label{eq:expectedfeature_noisycontexts} \end{align} is the expected feature map with respect to $P(c_t|\chat_t,\Hscr_{t-1,\chat})$.
%However, evaluating this is more challenging than Algorithm~\ref{alg:1}. Precisely, the predictive posterior distribution is obtained as $P(c_t|\chat_t,\Hscr_{t-1,\chat})=\Ebb_{P(\gammastar|\Hscr_{t-1,\chat})}[P(c_t|\chat_t,\gammastar)]$ by marginalizing the distribution $P(c_t|\chat_t,\gammastar)$, defined in \eqref{eq:c_chatgammastar},  over the posterior distribution $P(\gammastar|\Hscr_{t-1,\chat})$ of $\gammastar$. Differently from the denoising step of Algorithm~\ref{alg:1}, as the agent never observes true context $c_t$ at any iteration $t$, the posterior distribution $P(\gammastar|\Hscr_{t-1,\chat})$ must be evaluated solely based on the history of observed noisy contexts. As such, to obtain $P(\gammastar|\Hscr_{t-1,\chat})$, we use Bayes' theorem  with $P(\chat_t|\gammastar)$ defined in \eqref{eq:chat_gammastar} as the likelihood, \emph{i.e.},
%$P(\gammastar|\Hscr_{t-1,\chat}) \propto P(\gammastar) \prod_{\tau=1}^{t-1}P(\chat_{\tau}|\gammastar).$

\subsection{Linear-Gaussian Stochastic CBs} \label{sec:linearGaussianbandits} We now instantiate Algorithm~1 for Gaussian CBs. Specifically, we consider Gaussian bandits with the reward noise $\xi_t$ in \eqref{eq:rewardfunction} as Gaussian $\Nscr(0,\sigma^2)$ with mean $0$ and variance $\sigma^2>0$. We also assume a  Gaussian prior
 $P(\thetastar)=\Nscr( \boldsymbol{0},\lambda \Ibb)$  on the reward parameter $\thetastar$ with  mean zero and an $m \times m$ diagonal, covariance matrix with entries $\lambda>0$. Here, $\Ibb$ denotes the identity matrix.
We consider a multivariate Gaussian context distribution  $P(c)= \Nscr(\mu_c, \Sigma_c)$ with mean $\mu_c \in \mathbb{R}^d$ and covariance matrix $\Sigma_c \in \mathbb{R}^{d \times d}$. The context noise channel $P(\chat|c,\gammastar)$ is also similarly Gaussian with mean $(\gammastar+c)$ and covariance matrix $\Sigma_n\in \mathbb{R}^{d \times d}$. We assume the prior on noise channel parameter $\gammastar$ to be Gaussian  $P(\gammastar)=\Nscr( \boldsymbol{0},\Sigma_{\gamma})$ with $d$-dimensional zero mean vector $\boldsymbol{0}$ and covariance matrix $\Sigma_{\gamma} \in \mathbb{R}^{d \times d}.$ We assume that $\Sigma_c, \Sigma_{\gamma}$ and $\Sigma_n$ are all positive definite matrices known to the agent.

 For this setting, we can analytically evaluate the predictive posterior distribution $P(c_t|\chat_t,\Hscr_{t-1,\chat}) = \mathcal{N}(c_t|V_t,R_t^{-1})$ as a multi-variate Gaussian with inverse covariance matrix,
\begin{align}
     R_t = M-\Sigma_n^{-1}(H_t^{-1})^\top\Sigma_n^{-1}, \label{eq:variance_predcontext_denoising}\end{align} where $H_t= (t-1)\Sigma_n^{-1}-(t-2)\Sigma_n^{-1}M^{-1}\Sigma_n^{-1}+\Sigma_{\gamma}^{-1}$ and $M=\Sigma_c^{-1}+\Sigma_n^{-1}$, and with the mean vector \begin{align}
   \scalemath{0.96}{ \hspace{-0.1in} V_t = (R_t^{-1})^\top \Bigl(\Sigma_c^{-1}\mu_c+\Sigma_n^{-1}\chat_t-\Sigma_n^{-1}(H_t^{-1})^\top L_t^\top \Bigr), \hspace{-0.1in}}\label{eq:mean_predcontext_denoising}
\end{align} where $L_t^\top=\Sigma_n^{-1}M^{-1}(\Sigma_c^{-1}\mu_c+\Sigma_n^{-1}\chat_t)+(\Sigma_n^{-1}-\Sigma_n^{-1}M^{-1}\Sigma_n^{-1})\sum_{\tau=1}^{t-1}\chat_{\tau} -(t-1)\Sigma_n^{-1}M^{-1}\Sigma_c^{-1}\mu_c$. Derivations can be found in Appendix~\ref{app:posterior_predictive_nodelay}.

For the {\it modified-}TS step, we sample $\theta_t$ from a multi-variate Gaussian distribution $\bar{P}_t(\thetastar):=\bar{P}(\thetastar|\Hscr_{t-1,r,a,\chat})=\Nscr(\mu_{t-1}, \Sigma_{t-1}^{-1})$  whose inverse covariance matrix and mean respectively evaluate as
\begin{align}
\Sigma_{t-1} &= \frac{\Ibb}{\lambda}+\frac{1}{\sigma^2}\sum_{\tau=1}^{t-1}\psi(a_{\tau},\chat_{\tau}|\Hscr_{\chat})\psi(a_{\tau},\chat_{\tau}|\Hscr_{\chat}) ^\top\label{eq:variance_theta}\\
\mu_{t-1} &= \frac{\Sigma_{t-1}^{-1}}{\sigma^2} \Bigl(\sum_{\tau=1}^{t-1}r_{\tau}\psi(a_{\tau},\chat_{\tau}|\Hscr_{\chat})\Bigr), \label{eq:mean_theta}
\end{align} where $\psi(a_t,\chat_t|\Hscr_{\chat})$ is the expected feature map defined in \eqref{eq:expectedfeature_noisycontexts}. The  sampling distribution $\bar{P}(\thetastar|\Hscr_{t-1,r,a,\chat})$ considered above is different from the true posterior distribution \eqref{eq:trueposterior}, which is analytically intractable (see Appendix~\ref{app:analyticaltractability} for details). In Sec.~\ref{sec:bayesregret}, we show that the above choice of sampling distribution is indeed `close' to the true posterior.

%It is easy to see that \eqref{eq:variance_theta} and \eqref{eq:mean_theta} bear  resemblance to \eqref{eq:variance_theta_delayed} and \eqref{eq:mean_theta_delayed} used in Algorithm~\ref{alg:1} in that the feature map $\phi(a_{\tau},c_{\tau})$ therein is replaced  by the expected feature map $\psi(a_{\tau},\chat_{\tau}|\Hscr_{\chat})$. We use the expected feature map in Algorithm~\ref{alg:2} since the agent cannot evaluate $\phi(a_{\tau},c_{\tau})$ as it never observes the true context $c_{\tau}$. 
 
%{\color{red}\begin{remark} The choice of the Gaussian framework above is due to the easy tractability of posterior and predictive posterior distributions involved in the TS algorithm. We note that similar Gaussian contextual bandit problem with Gaussian context noise has been studied in \citep{kim2023contextual} wherein they developed an UCB-algorithm that achieves sub-linear frequentist regret.
%\end{remark}}
\subsection{Bayesian Regret Analysis}
In this section, we derive information-theoertic upper bounds on the Bayesian regret \eqref{eq:Bayesian_cumregret} of the modified TS algorithm for Gaussian CBs. To this end, we first outline the key information-theoretic tools required to derive our bound.
\subsubsection{Preliminaries}
 To start, let $P(x)$ and $Q(x)$ denote two probability distributions defined over the space $\Xscr$ of random variables $x$.  Then, the Kullback Leibler (KL)-divergence between the distributions $P(x)$ and $Q(x)$ is defined as
\begin{align}
\scalemath{0.96}{D_{\rm KL}(P(x)||Q(x))=\Ebb_{P(x)}\biggl[\log \frac{P(x)}{Q(x)}\biggr]},
\end{align}if $P(x)$ is absolutely continuous with respect to $Q(x)$, and takes value $\infty$ otherwise. If $x$ and $y$ denote two random variables described by the joint probability distribution $P(x,y)$, the mutual information $I(x;y)$ between $x$ and $y$ is defined as $I(x;y)=D_{\rm KL}(P(x,y) \Vert P(x)P(y))$,  where $P(x)$ (and $P(y)$) is the marginal distribution of $x$ (and $y$). More generally, for three random variables $x$, $y$ and $z$ with joint distribution $P(x,y,z)$, the conditional mutual information $I(x;y|z)$ between $x$ and $y$ given $z$ evaluates as $I(x;y|z) = \Ebb_{P(z)}[D_{KL}(P(x,y|z) \Vert P(x|z)P(y|z))]$ where $P(x|z)$ and $P(y|z)$ are conditional distributions.
We will also use the following variational representation of the KL-divergence, also termed the {\it Donskar-Varadhan (DV)} inequality,
\begin{align}
\scalemath{0.95}{
\hspace{-3.5 mm}D_{\rm KL}(P(x) \Vert Q(x)) \hspace{-1 mm}\geq \Ebb_{P(x)}[f(x)] \hspace{-1 mm}-\hspace{-1 mm}\log \Ebb_{Q(x)}[\exp(f(x))], \label{eq:DV_inequality}}
\end{align}which holds for any measurable function $f:\mathcal{X} \rightarrow \mathbb{R}$ satifying the inequality $\Ebb_{Q(x)}[\exp(f(x))]<\infty$. 
%In particular, we note that if $f(x) \sim \Nscr(\mu,\sigma^2)$, we have that for any $\lambda \in \mathbb{R}$, $\Ebb[\exp\lambda(f(x))] =\exp(\lambda \mu +\lambda^2 \sigma^2/2)$. 

\subsubsection{Information-Theoretic Bayesian Regret Bounds} \label{sec:bayesregret}
To analyze the Bayesian regret of the proposed algorithm for Gaussian contextual bandits, we start by defining  \begin{align}\hat{a}_t =\arg \max_{a \in \Acal} \psi(a,\chat_t|\Hscr_{\chat})^\top\thetastar \label{eq:interm_action} \end{align} as the action that maximizes the mean reward $\psi(a,\chat_t|\Hscr_{\chat})^\top\thetastar$ corresponding to reward parameter $\thetastar$.
%We also use the notation $\mathcal{F}_t=\Hscr_{t-1,r,a,\chat} \cup \chat_t$ to denote the set of all observations until the action selection at iteration $t$.
Using the above, the Bayesian cumulative regret \eqref{eq:Bayesian_cumregret} for the proposed TS algorithm $\pi^{\rm{TS}}$ can be decomposed as 
\begin{align}
&\Rcal^T(\pi^{\rm{TS}})=\Rcal^T_{{\rm CB}}+ \Rcal^T_{{\rm EE1}}+ \Rcal^T_{{\rm EE2}}, \label{eq:Bayesregret_decompsition} \hspace{0.2cm} \mbox{where} \\
&\scalemath{0.96}{ \Rcal^T_{{\rm CB}} =\sum_{t=1}^T\mathbb{E}\Bigl[\psi(\hat{a}_t,\chat_t|\Hscr_{\chat})^\top\thetastar-\psi(a_t,\chat_t|\Hscr_{\chat})^\top\thetastar   \Bigr]},\nonumber \\ & \scalemath{0.96}{\Rcal^T_{\rm EE1}= \sum_{t=1}^T \Ebb \Bigl[ \psi(a^{*}_t,\chat_t|\gammastar)^\top \thetastar -\psi(\hat{a}_t,\chat_t|\Hscr_{\chat})^\top\thetastar \Bigr],} \nonumber \\& \scalemath{0.96}{\Rcal^T_{\rm EE2}= \sum_{t=1}^T \Ebb\Bigl[\psi(a_t,\chat_t|\Hscr_{\chat})^\top\thetastar-\psi(a_t,\chat_t|\gammastar)^\top \thetastar \Bigr] \nonumber. }
\end{align}

In \eqref{eq:Bayesregret_decompsition}, the first term $\Rcal^T_{{\rm CB}}$ quantifies the Bayesian regret of our action policy  \eqref{eq:actionpolicy_nocontexts} with respect to the action policy \eqref{eq:interm_action} for a CB with mean reward function $\psi(a,\chat_t|\Hscr_{\chat})^\top\thetastar$. The second term $\Rcal^T_{{\rm EE1}}$ accounts for the average difference in the cumulative mean rewards of the oracle optimal action policy \eqref{eq:oracle_action}, evaluated using the exact predictive distribution $P(c_t|\chat_t,\gammastar)$, and our action policy  \eqref{eq:actionpolicy_nocontexts}, that uses the inferred predictive posterior distribution $P(c_t|\chat_t,\Hscr_{t-1,\chat})$. In this sense,  $\Rcal^T_{{\rm EE1}}$ captures the error in approximating the exact predictive distribution $P(c_t|\chat_t,\gammastar)$ via the inferred predictive distribution $P(c_t|\chat_t,\Hscr_{\chat})$.  The third term $\Rcal^T_{{\rm EE2}} $ similarly accounts for the average approximation error. 
To derive an upper bound on the Bayesian regret $\mathcal{R}^T(\pi^{\rm TS})$, we separately upper bound each of the three terms in \eqref{eq:Bayesregret_decompsition}. To this end, we make the following assumption.
\begin{assumption}\label{assum:1}
 The feature map $\phi(\cdot,\cdot) \in \mathbb{R}^m$ has bounded norm, i.e., $\Vert \phi(\cdot,\cdot)\Vert_2 \leq 1$.
\end{assumption} 
The following lemma presents an  upper bound on  $\Rcal^T_{{\rm CB}}$.
%However, to evaluate this term, we cannot directly leverage the information-theoretic approach of \citep{neu2022lifting} as we did for Algorithm~\ref{alg:1} since, as pointed out earlier, our algorithm samples $\theta_t$ from a distribution that is different from the true posterior $P_t(\thetastar)$. Accordingly, the term $\Rcal^T_{{\rm CB}}$  also account for the {\it posterior mismatch} between the true posterior $P_t(\thetastar)$ and the sampling distribution $\bar{P}_t(\thetastar)$.
%The following lemma presents an upper bound on $\Rcal^T_{{\rm CB}}$. 
\begin{lemma}\label{lem:r_CB_withcontextdistributions}
%Let $\Sigma_c =\sigma^2_c \Ibb$, $\Sigma_n=\sigma^2_n \Ibb$ and $\sigma_c =\sigma^2_c \Ibb$ where $\Ibb$ denotes the identity matrix. Assume that the feature map $\phi(a,c)=G(a)c$ where $G(a)$ is a $d \times d$ matrix such that satisfies Assumption~\ref{assum:1}.  
Under Assumption~\ref{assum:1}, the following upper bound holds if $\frac{\lambda}{\sigma^2}\leq \frac{d}{T} \leq 1$, 
\begin{align}
   & \scalemath{0.9}{\Rcal^T_{{\rm CB}}\hspace{-0.05cm}\leq \hspace{-0.05cm}U(m,\frac{d\sigma^2}{T})\hspace{-1.5 mm} + \hspace{-1.5 mm} \sqrt{2d\sigma^2  \sum_{t=1}^T D_t}}  \scalemath{0.9}{+ \sqrt{2d\sigma^2 \Bigl(T \log(K)+\sum_{t=1}^T  D_t\Bigr)}},
    \label{eq:RCB_ub_nodelay}
\end{align}

where $D_t =\Ebb[D_{\rm KL}(P_t(\thetastar)\Vert \bar{P}_t(\thetastar))] $  and 
\begin{align}
\hspace{-3 mm}\scalemath{0.9}{U(m,\lambda)} &\scalemath{0.9}{=\hspace{-1 mm}\sqrt{2Tm\sigma^2\min\{m, 2+2\log K\}\log \Bigl(1+\frac{T\lambda}{m\sigma^2}\Bigr)}}\label{eq:UCB}.
\end{align}

In particular, if the feature map $\phi(a,c)=G(a)c$ with $G(a)$ being a $d \times d$ matrix  satisfying Assumption~\ref{assum:1} with $m=d$, 
\begin{align}
   \scalemath{0.9}{ \sum_{t=1}^T D_t \leq \frac{dT}{4}.} \label{eq:KL_bound}
\end{align}
\end{lemma}

To derive the upper bound in \eqref{eq:RCB_ub_nodelay}, we leverage results from \cite{neu2022lifting} that studies information-theoretic Bayesian regret of standard contextual TS algorithms. However, the results do not directly apply to our algorithm due to the \textit{posterior mismatch} between the sampling distribution $\bar{P}_t(\thetastar)$ and the true posterior distribution $P_t(\thetastar)$. Consequently, our upper bound \eqref{eq:RCB_ub_nodelay} consists of three terms: the first term, defined as in \eqref{eq:UCB}, corresponds to the upper bound on the Bayesian regret of contextual TS that assumes $\bar{P}_t(\thetastar)$ as the true posterior. This can be obtained by applying \cite[Cor.~2]{neu2022lifting}.  The second and third terms account for the posterior mismatch via the expected KL-divergence $D_t$ between the true posterior $P_t(\thetastar)$ and the sampling distribution $\bar{P}_t(\thetastar)$.
%In contrast to standard contextual TS algorithm, our proposed TS The derivation of the information-theoretic upper bound in \eqref{eq:RCB_ub_nodelay}  is to account for the posterior mismatch between the true posterior distribution and the sampling distribution. While information-theoretic upper bounds for contextual TS algorithms exist \cite{neu2022lifting}, they have no posterior mismatch. Consequently, to derive the required upper bound, we decompose $\mathcal{R}^T_{\rm CB}$ into a Bayesian regret term that assumes the sampling distribution to tbe the true distribution, and a term that can account for posterior mismatch. Accordingly, the first term in \eqref{eq:RCB_ub_nodelay} corresponds to an upper bound on the Bayesian regret of standard TS for a CB with $\bar{P}_t(\thetastar)$ as the true posterior distribution. The second and third terms  quantify the posterior mismatch via the expected KL divergence between the posterior $P_t(\thetastar)$ and the sampling distribution $\bar{P}_t(\thetastar)$. 
In particular, when the feature map $\phi(a,c)$  is linear in $c$, the above expected KL divergence can be bounded as in \eqref{eq:KL_bound} provided that the prior $P(\thetastar)=\Nscr(\boldsymbol{0},\lambda\Ibb)$ is sufficiently concentrated.  This ensures that the contribution of posterior mismatch to the Bayes regret scales as $O(d\sqrt{T})$.

The following lemma gives an upper bound on the sum $\Rcal^T_{{\rm EE1}}+\Rcal^T_{{\rm EE2}}$.
%As explained before, these terms account for the error resulting from approximating the exact predictive distribution $P(c_t|\chat_t,\gammastar)$ with the inferred posterior predictive distribution $P(c_t|\chat_t,\Hscr_{t-1,\chat}).$
\begin{lemma}\label{lem:estmationerror_nocontext}
    Under Assumption~\ref{assum:1}, the following upper bound holds for $\delta \in (0,1)$  if $\frac{\lambda}{\sigma^2}\leq \frac{d}{T} \leq 1$,
    \begin{align}
        & \scalemath{0.96}{\Rcal^T_{\rm EE1}+\Rcal^T_{\rm EE2} \leq 2 \Rcal^T_{\rm EE1} \leq 4\delta^2 \sqrt{\frac{md\sigma^2 T}{2\pi}}}\non \\
        & \scalemath{0.96}{+ 2\sqrt{4md\sigma^2 \log \Bigl(\frac{2m}{\delta} \Bigr) \sum_{t=1}^T I(\gammastar;c_t|\chat_t,\Hscr_{t-1,\chat})}}. 
    \end{align} In particular, if $\phi(a,c)=G(a)c \in \mathbb{R}^d$ satisfies Assumption~\ref{assum:1} with $m=d$,  $\Sigma_c=\sigma^2_c\Ibb$, $\Sigma_n =\sigma^2_n \Ibb$, and $\Sigma_{\gamma} =\sigma_{\gamma}^2\Ibb$, where $\sigma^2_n,\sigma^2_c,\sigma^2_{\gamma}>0$, we have the tigher bound of
    \begin{align}
    \scalemath{0.96}{\Rcal^T_{{\rm EE1}}+ \Rcal^T_{{\rm EE2}}\leq   2\sqrt{2L \sum_{t=1}^T I(\gammastar;c_t|\chat,\Hscr_{t-1,\chat})}},
    \end{align} where $L=dK\nu \Bigl(  \sigma^2_n + \frac{\sigma^2_c \sigma^2_{\gamma}}{T(\sigma^2_c+\sigma^2_n)}+\sigma^2_c \frac{\log(T-1)}{T}\Bigr)$ \footnote{$\nu=\sigma^2 \sigma^2_c/(\sigma^2_c+\sigma^2_n)\max_a {\rm Tr}(G(a)^\top G(a))$}, with
    \begin{align}
    \scalemath{0.96}{\sum_{t=1}^T I(\gammastar;c_t|\chat_t,\Hscr_{t-1,\chat}) \leq \frac{d\sigma^2_c}{2\sigma^2_n} \Bigl(\frac{\sigma^2_{\gamma}}{\sigma^2_c+\sigma^2_n}+ \log (T-1)\Bigr)}\non .
    \end{align}
\end{lemma}
Lemma~\ref{lem:estmationerror_nocontext} shows that the error in approximating $P(c_t|\chat_t,\gammastar)$ with $P(c_t|\chat_t,\Hscr_{t-1,\chat})$, on average, can be quantified via the conditional mutual information $I(\gammastar;c_t|\chat_t,\Hscr_{t-1,\chat})$ between $\gammastar$ and true context $c_t$ given knowledge of observed noisy contexts upto and including iteration $t$. 
%A general upper bound that does not require the assumption of linear feature map is provided in Appendix~\ref{app:generalupperbound}.
%
Combining Lemma~\ref{lem:r_CB_withcontextdistributions} and Lemma~\ref{lem:estmationerror_nocontext} gives us the following upper bound on $\Rcal^T(\pi^{\rm{TS}})$.
\begin{theorem} \label{thm:Bayesianregret_nodelay} Assume that $\phi(a,c)=G(a)c \in \mathbb{R}^d$,  $\Sigma_c=\sigma^2_c\Ibb$, $\Sigma_n =\sigma^2_n \Ibb$, and $\Sigma_{\gamma} =\sigma_{\gamma}^2\Ibb$, where $\sigma^2_n,\sigma^2_c,\sigma^2_{\gamma}>0$.  If $\frac{\lambda}{\sigma^2}\leq \frac{d}{T} \leq 1$, we have the following upper bound,
\begin{align}
    &\scalemath{0.96}{\Rcal^T(\pi^{\rm{TS}}) \leq U(d,d\sigma^2/T)+\sqrt{d\sigma^2 \Bigl(2T \log(K)+\frac{dT}{2}\Bigr)}} \nonumber \\&\scalemath{0.9}{+\sqrt{\frac{d^2T\sigma^2}{2} } +2\sqrt{2L\frac{d\sigma^2_c}{\sigma^2_n} \Bigl(\frac{\sigma^2_{\gamma}}{\sigma^2_n+\sigma^2_c}+ \log (T-1)\Bigr)}},\non
\end{align}where $L$ is as defined in Lemma~\ref{lem:estmationerror_nocontext}.
\end{theorem}

The theorem above shows that the proposed TS algorithm achieves $O(d\sqrt{T})$ regret for feature maps linear in context vector when the prior $P(\thetastar)$ is highly informative.
\section{TS for CB with Delayed True Contexts}\label{sec:TS_delayed}
In this section, we consider the CBs with delayed true context setting where the agent observes the true context $c_t$ after it observes the reward $r_t$ corresponding to the chosen action $a_t$. Note that at the time of choosing action $a_t$, the agent has access only to noisy contexts. We specialize our TS algorithm to this setting, and call it Algorithm~2 (or $\pi^{\rm TS}_{\rm delay})$.
\begin{comment}
\begin{algorithm}[h!]
\caption{\textbf{Algorithm 2}: TS with Delayed Contexts ($\pi^{\rm{TS}}_{\rm delay}$)}\label{alg:1}
\begin{algorithmic}[1]
\For{$t=1,\hdots,T$}
\State The environment selects a true context $c_t$.
\State Agent observes noisy context $\chat_t$.
\State Agent evaluates the predictive posterior distribution $P(c_t|\chat_t,\Hscr_{t-1,c,\chat})$ as in \eqref{eq:predictiveposterior_delayed}.
\State Agent samples $\theta_t \sim {P}(\thetastar|\Hscr_{t-1,r,a,c})$
\State Agent chooses action $a_t$ as in \eqref{eq:actionpolicy_nocontexts}.
\State Agent observes reward $r_t$ corresponding to $a_t$, as well as the true context $c_t$.
%\State 7. Update the history $\Hscr_{t,r,a,c,\chat}=\Hscr_{t-1,r,a,c,\chat} \cup (r_t,a_t,c_t,\chat_t)$
\EndFor
\end{algorithmic}
\end{algorithm}
\end{comment}

Algorithm~2 follows similar steps as in Algorithm~1. However, different from Algorithm~1, at $t$th iteration, the agent knows the history $\Hscr_{t-1,c,\chat}$ of true contexts in addition to that of noisy contexts. Consequently, in the \textit{de-noising} step, the agent
evaluates the predictive posterior distribution as
\begin{align}
    P(c_t|\chat_t,\Hscr_{t-1,c,\chat})=\Ebb_{P(\gammastar|\Hscr_{t-1,c,\chat})}[P(c_t|\chat_t,\gammastar)], \label{eq:predictiveposterior_delayed} 
\end{align}where $P(c_t|\chat_t,\gammastar)$ is as defined in \eqref{eq:c_chatgammastar} and posterior distribution $P(\gammastar|\Hscr_{t-1,c,\chat})$ is obtained via Baye's rule as
$ P(\gammastar|\Hscr_{t-1,c,\chat}) \propto P(\gammastar) \prod_{\tau=1}^{t-1}P(c_{\tau},\chat_{\tau}|\gammastar)$ using the history of true and noisy contexts.

For Gaussian context noise as considered in Sec.\ref{sec:linearGaussianbandits}, 
%The proposed algorithm described in Algorithm~\ref{alg:1} implements two steps in each iteration $t \in \mathbb{N}$.  The first step, called the {\it denoising step}, uses the current observed noisy context $\chat_t$, and the history $\Hscr_{t-1,c,\chat}=\{c_{\tau},\chat_{\tau}\}_{\tau=1}^{t-1}$ of past observed noisy contexts and revealed true contexts, to obtain a {\it predictive posterior distribution} $P(c_t|\chat_t,\Hscr_{t-1,c,\chat})$ of the true context. This is a two-step process where firstly, we use the history $\Hscr_{t-1,c,\chat}$ of observations to update the agent's belief about the unknown noise channel parameter $\gammastar$ to a posterior distribution $P(\gammastar|\Hscr_{t-1,c,\chat})$. Thanks to the agent's knowledge of the prior $P(\gammastar)$, the context distribution $P(c)$ as well as the noise channel likelihood $P(\chat_t|c_t,\gammastar)$, evaluating the posterior distribution is a consequence of applying the Baye's rule. The predictive posterior distribution is then obtained as $P(c_t|\chat_t,\Hscr_{t-1,c,\chat})=\Ebb_{P(\gammastar|\Hscr_{t-1,c,\chat})}[P(c_t|\chat_t,\gammastar)]$, where $P(c_t|\chat_t,\gammastar)$ is as defined in \eqref{eq:c_chatgammastar}. 
 the predictive posterior distribution $P(c_t|\chat_t,\Hscr_{t-1,c,\chat})=\Nscr(\tilde{V}_t,\tilde{R}_t^{-1})$ is  multivariate Gaussian with the inverse covariance matrix,
\begin{align}
\tilde{R_t}& =M-\Sigma_n^{-1}\tilde{H}_t^{-1}\Sigma_n^{-1} \label{eq:covariance_predictedcontext_delayed},
\end{align} and the mean vector 
\begin{align}
       \tilde{V_t}& = \tilde{R}_t^{-1}\Bigl( \Sigma_c^{-1}\mu_c+\Sigma_n^{-1}\chat_t+ \Sigma_n^{-1}\tilde{H}_t^{-1}\Sigma_n^{-1}\sum_{\tau=1}^{t-1}(\chat_{\tau}-c_{\tau})\nonumber \\& - \Sigma_n^{-1}\tilde{H}_t^{-1}\Sigma_n^{-1}M^{-1}(\Sigma_c^{-1}\mu_c-\Sigma_n^{-1}\chat_t)\Bigr) \label{eq:mean_predictedcontext_delayed},
\end{align} where $M =\Sigma_c^{-1}+\Sigma_n^{-1}$ and $\tilde{H}_t=\Sigma_n^{-1}M^{-1}\Sigma_n^{-1}+(t-1)\Sigma_n^{-1}+\Sigma_{\gamma}^{-1}$. Derivation can be found in App.~\ref{app:derivations_delayed}.

Following the denoising step, the next step in Algorithm~2 is a conventional Thompson sampling step, thanks to access to delayed true contexts. Consequently, the agent can evaluate the posterior distribution $P(\thetastar|\Hscr_{t-1,r,a,c})$ with known contexts  as in \eqref{eq:posterior} and use it to sample $\theta_t \sim P(\thetastar|\Hscr_{t-1,r,a,c})$. For  Gaussian bandit with Gaussian prior on $\thetastar$, the posterior distribution $P(\thetastar|\Hscr_{t-1,r,a,c})=\Nscr(\tilde{\mu}_{t-1},\tilde{\Sigma}_{t-1}^{-1})$ is a multivariate Gaussian distribution whose inverse covariance matrix and mean respectively evaluate as
\begin{align}
\tilde{\Sigma}_{t-1} &= \frac{1}{\lambda}\Ibb+\frac{1}{\sigma^2}\sum_{\tau=1}^{t-1}\phi(a_{\tau},c_{\tau})\phi(a_{\tau},c_{\tau})^\top\label{eq:variance_theta_delayed}\\
\tilde{\mu}_{t-1} &= \frac{\tilde{\Sigma}_{t-1}^{-1}}{\sigma^2} \Bigl(\sum_{\tau=1}^{t-1}r_{\tau} \phi(a_{\tau},c_{\tau})\Bigr). \label{eq:mean_theta_delayed}
\end{align}
Using the sampled $\theta_t$ and the obtained predictive posterior distribution $P(c_t|\chat_t,\Hscr_{t-1,c,\chat})$, the agent then chooses action $a_t$ as
\begin{align}
a_t
%&= \arg \max_{a \in \Acal}\Ebb_{P(c_t|\chat_t,\Hscr_{t-1,c,\chat})}[\phi(a,c_t)^\top \theta_t] \nonumber \\
&= \arg \max_{a \in \Acal} \psi(a,\chat_t|\Hscr_{c,\chat})^\top\theta_t, \label{eq:action_TS_delayed}
\end{align}where we use the expected feature map $\psi(a_t,\chat_t|\Hscr_{c,\chat}):=\Ebb_{P(c_t|\chat_t,\Hscr_{t-1,c,\chat})}[\phi(a_t,c_t)]$.
\subsection{Information-Theoretic Bayesian Regret Bounds}\label{sec:regretanalysis_delayed}
In this section, we derive an information-theoretic upper bound on the Bayesian regret \eqref{eq:Bayesian_cumregret} of Algorithm~2 for Gaussian CBs. To prevent overloading notation,  we re-use 
\begin{align}
    \hat{a}_t = \arg \max_{a \in \Acal}\psi(a,\chat_t|\Hscr_{c,\chat})^\top \thetastar \label{eq:optimalaction_thetastar}
\end{align} to define the optimal action maximizing the mean reward $\psi(a,\chat_t|\Hscr_{c,\chat})^\top \thetastar$.
%This can be interpreted as the optimal action taken by the agent had it known the reward parameter $\thetastar$.
%Throughout this section, we use $\Fcal_t =\Hscr_{t-1,r,a,c,\chat}\cup \chat_t$ to denote the set of all observations until the action selection at iteration $t$, and use it to define the conditional expectation $\Ebb_t[\cdot]:=\Ebb[\cdot|\Fcal_t]$ of $`\cdot$' with respect to the observations $\Fcal_t$.

To derive the upper bound,  we first decompose the Bayesian cumulative regret \eqref{eq:Bayesian_cumregret} of Algorithm~2 ($\pi^{\rm{TS}}_{\rm delay}$), similar to \eqref{eq:Bayesregret_decompsition}, into the following three terms, 
\begin{align}
&\scalemath{0.96}{\Rcal^T(\pi^{\rm{TS}}_{\rm delay}) = \Rcal^T_{\rm d,CB}+\Rcal^T_{\rm d,EE1} +\Rcal^T_{\rm d,EE2} \hspace{0.2cm} \mbox{where}}, \label{eq:Bayesregret_decompsition_delayed} \\ &\scalemath{0.96}{\Rcal^T_{\rm d,CB}= \sum_{t=1}^T\mathbb{E}\Bigl[\psi(\hat{a}_t,\chat_t|\Hscr_{c,\chat})^\top\thetastar-\psi(a_t,\chat_t|\Hscr_{c,\chat})^\top\thetastar   \Bigr]},\nonumber \\ &\scalemath{0.96}{\Rcal^T_{\rm d,EE1}=\sum_{t=1}^T \Ebb \Bigl[ \psi(a^{*}_t,\chat_t|\gammastar)^\top \thetastar -\psi(\hat{a}_t,\chat_t|\Hscr_{c,\chat})^\top\thetastar \Bigr]}, \nonumber\\ &\scalemath{0.96}{\Rcal^T_{\rm d,EE2}= \sum_{t=1}^T \Ebb\Bigl[\psi(a_t,\chat_t|\Hscr_{c,\chat})^\top\thetastar-\psi(a_t,\chat_t|\gammastar)^\top \thetastar \Bigr]}. \nonumber
\end{align} An upper bound on  $\Rcal^T(\pi^{\rm{TS}}_{\rm delay})$ can be obtained by separately bound each of the three terms in \eqref{eq:Bayesregret_decompsition_delayed}. 

In \eqref{eq:Bayesregret_decompsition_delayed}, the first term $\Rcal^T_{\rm d,CB}$ corresponds to the Bayesian cumulative regret of a standard contextual TS algorithm that uses $\psi(a,\chat_t|\Hscr_{c,\chat})^\top\thetastar$ for $a \in \Acal$ as the mean reward function. Note that due to availability of delayed true contexts, there is no posterior mismatch in Algorithm~2. Hence, we apply \cite[Cor.~3]{neu2022lifting} to yield the following upper bound on $\Rcal^T_{\rm d,CB}$.
%\begin{assumption}\label{assum:1}
 %The feature map $\phi(\cdot,\cdot) \in \mathbb{R}^m$ has bounded norm, i.e., $\Vert \phi(\cdot,\cdot)\Vert_2 \leq 1$.
%\end{assumption}
\begin{lemma}\label{lem:R_CB_delayed}
Under Assumption~\ref{assum:1}, the following upper bound on $\Rcal^T_{\rm d,CB}$ holds for $\frac{\lambda}{\sigma^2} \leq 1$,
\begin{align}
    \Rcal^T_{\rm d,CB} &\leq U(m,\lambda),\label{eq:RCB_ub}  
\end{align} where $U(m,\lambda)$ is defined as in \eqref{eq:UCB}.
\end{lemma}

Lemma~\ref{lem:R_CB_delayed} gives a tighter bound in comparison to Lemma~\ref{lem:r_CB_withcontextdistributions} where the posterior mismatch results in additional error terms in the regret bound.

We now upper bound the second term $\Rcal^T_{\rm d,EE1}$  of \eqref{eq:Bayesregret_decompsition_delayed}, which similar  to the term $\Rcal^T_{\rm EE1}$  in \eqref{eq:Bayesregret_decompsition},  captures the error in approximating the exact predictive distribution $P(c_t|\chat_t,\gammastar)$ via the inferred predictive distribution $P(c_t|\chat_t,\Hscr_{c,\chat})$.  The following lemma shows that this approximation error over $T$ iterations can be quantified, on average, via the mutual information $I(\gammastar;\Hscr_{T,c,\chat})$ between $\gammastar$ and the $T$-length history of observed true and noisy contexts. This bound also holds for the third term $\Rcal^T_{\rm d,EE2}$ of \eqref{eq:Bayesregret_decompsition_delayed} which similarly accounts for the average approximation error.
%The following lemma thus presents an upper bound on the sum $\Rcal^T_{\rm d,EE1}+\Rcal^T_{\rm d,EE2}$.
%\begin{figure*}[h]
%\centering
 %   \subfloat[]{\includegraphics[width=0.3\linewidth]{synresults_13Oct_6_plotalone.png}}\label{fig1a}
 %   \subfloat[]{\includegraphics[width=0.3\linewidth]{synresults_24Jan_logistics_v5.4_plotalone.png}}\label{fig1b}
  %  \subfloat[]{\includegraphics[width=0.3\linewidth]{movielens_v5.1_plotalone.png}}\label{fig1c}\
  %  \subfloat[Sub-Caption d]{\fbox{\includegraphics[width=0.47\linewidth]{figure2}}\label{fig1d}}\\
 %   \subfloat[Sub-Caption e]{\fbox{\includegraphics[width=0.47\linewidth]{figure1}}\label{fig1e}}\
  %  \subfloat[Sub-Caption f]{\fbox{\includegraphics[width=0.47\linewidth]{figure2}}\label{fig1f}}
%\caption{caption here}\label{fig1}
%\end{figure*}
\begin{figure*}[t]
    \centering
    \begin{minipage}{0.33\textwidth}
        \centering
\includegraphics[scale=0.4, clip=true, trim= 0.2in 0in 0in 0.35in]{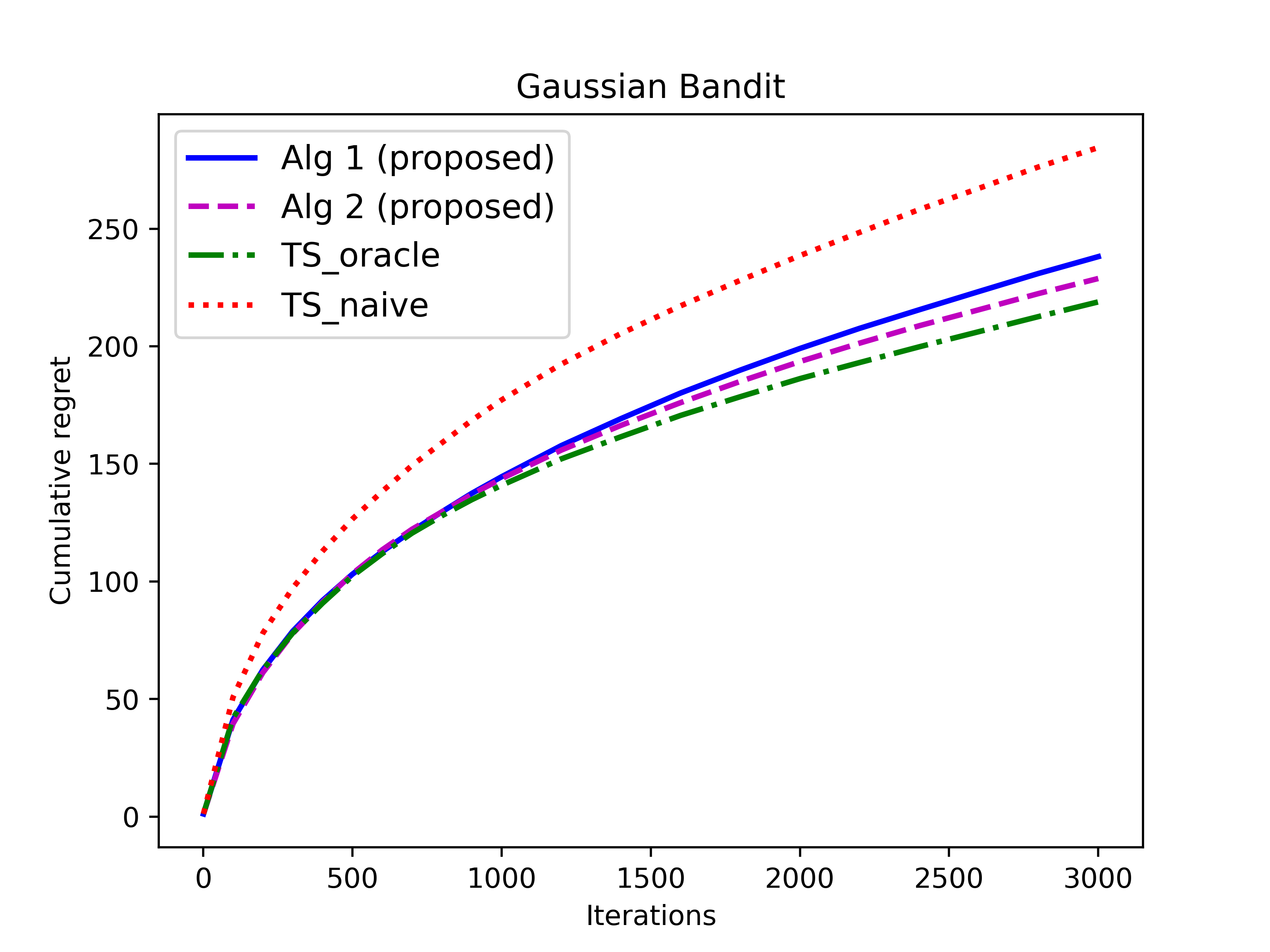} % first figure itself
    \end{minipage}\hfill
    \begin{minipage}{0.33\textwidth}
        \centering
\includegraphics[scale=0.4, clip=true, trim= 0.39in 0in 0in 0.3in]{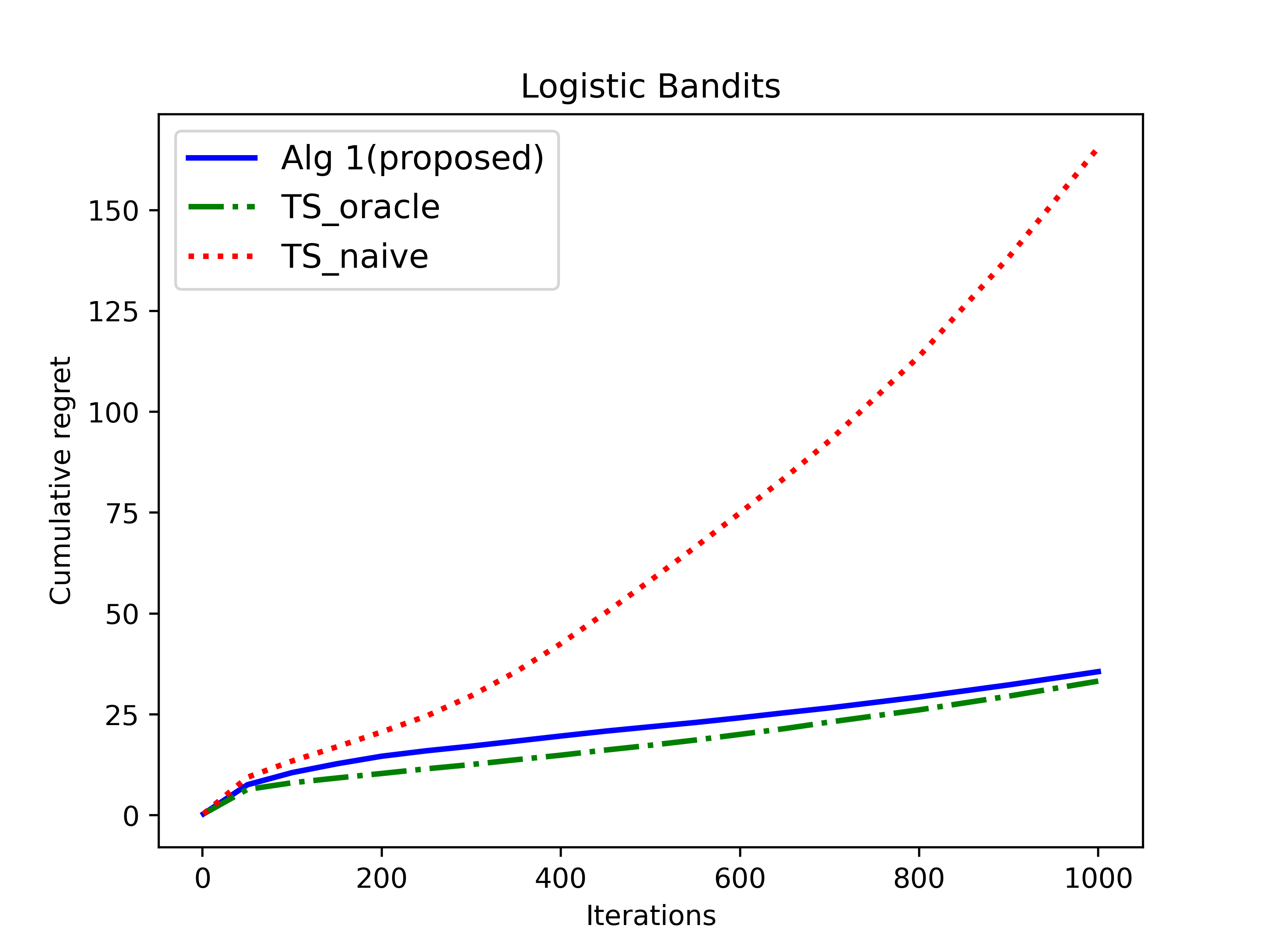} % second figure itsel
    \end{minipage}\hfill
    \begin{minipage}{0.33\textwidth}
        \centering
        \includegraphics[scale=0.4, clip=true, trim= 0.3in 0.1in 0in 0.3in]{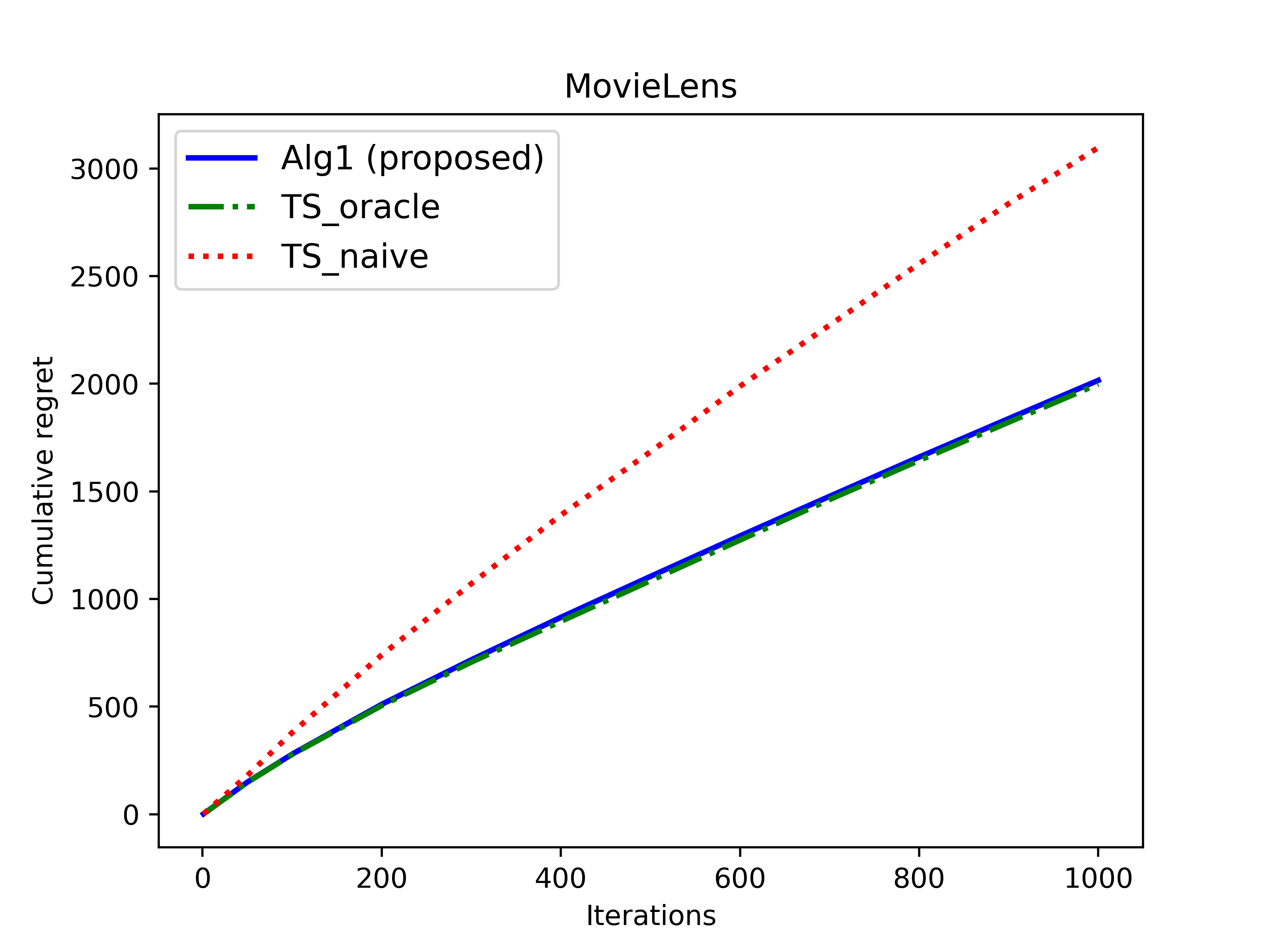} % third figure itsel
    \end{minipage}
    \caption{\small{{Comparison of Bayesian regret of proposed algorithms with baselines as a function of number of iterations. {\it (Left)}: Gaussian bandits with  $K=40$, $\sigma^2_n=\sigma^2_{\gamma}=1.1$; {\it (Center)} Logistic bandits with $K=40$, $\sigma^2_n=2$, $\sigma^2_{\gamma}=2.5$; {\it (Right)} MovieLens dataset with added Gaussian context noise and Gaussian prior: parameters set as $\sigma^2_n=0.1$, $\sigma^2_{\gamma}=0.6$.}}} \label{fig:expt1}
\end{figure*}
\begin{lemma}\label{lem:EE_1}
Under Assumption \ref{assum:1}, for any $\delta \in (0,1)$, we have the following upper bound,
\begin{align}
     &\scalemath{0.96}{\Rcal^T_{{\rm d,EE1}} +\Rcal^T_{{\rm d,EE2}}\leq 2 \Rcal^T_{{\rm d,EE1}}} \non \\& 
   \scalemath{0.96}{ \leq  4\sqrt{m\lambda T \log\Bigl(\frac{2m}{\delta}\Bigr)I(\gammastar;\Hscr_{T,c,\chat})} +2T \delta^2 \sqrt{\frac{2m\lambda}{\pi}}}. \non\end{align}
     Furthermore, if $\Sigma_{\gamma}= \sigma_{\gamma}^2 \Ibb$ and $\Sigma_n =\sigma^2_n \Ibb$ for $\sigma_{\gamma}^2, \sigma_n^2>0$,
     %we have
     \begin{align}
\scalemath{0.96}{ I(\gammastar;\Hscr_{T,c,\chat})=\frac{d}{2}\log \Bigl(1+(T-1)\frac{\sigma^2_{\gamma}}{\sigma^2_n} \non\Bigr)}.
\end{align}
\end{lemma}

Combining Lemma~\ref{lem:R_CB_delayed} and Lemma~\ref{lem:EE_1} then gives us the upper bound on $\Rcal^T(\pi^{\rm TS}_{\rm delay})$.
\begin{theorem}\label{thm:bayesianregret_delayedcontext}
   Under  Assumption \ref{assum:1} and with parameters $\Sigma_{\gamma}= \sigma_{\gamma}^2 \Ibb$,  $\Sigma_n =\sigma^2_n \Ibb$ for $\sigma_{\gamma}^2, \sigma_n^2>0$, the following inequality holds for $\delta \in (0,1)$
   \begin{align}
       &\scalemath{0.96}{\Rcal^T(\pi^{\rm TS}_{\rm delay}) \leq U(m,\lambda)+ 2T \delta^2 \sqrt{\frac{2m\lambda}{\pi}}} \non \\&\scalemath{0.96}{+2\sqrt{2\lambda m Td\log \Bigl( 1+(T-1)\frac{\sigma^2_{\gamma}}{\sigma^2_n}\Bigr)  \log \Bigl(\frac{2m}{\delta} \Bigr) }}.
   \end{align}
\end{theorem}
Theorem~\ref{thm:bayesianregret_delayedcontext} shows that Algorithm~2 achieves $O(\sqrt{Tm \log(T)})$ regret with the choice of $\delta =1/T$ if $m >2(1+\log K)$. Furthermore, due to the absence of posterior mistmatch, the upper bound above is tighter than that of Theorem~\ref{thm:Bayesianregret_nodelay}.

\section{Experiments and Final Remarks}

In this section, we experimentally validate the performance of the proposed algorithms on synthetic and real-world datasets. Details of implementation can be found in App.~\ref{app:experiments}.

\textbf{Synthetic Datasets}: For synthetic datasets, we go beyond Gaussian bandits and evaluate our algorithms for logistic contextual bandits (see Fig.~\ref{fig:expt1}(Left) and (Center)). In both these settings, we consider Gaussian contexts and context noise as in Sec.~\ref{sec:linearGaussianbandits} with parameters $\Sigma_c =\Ibb$, $\Sigma_{\gamma}=\sigma^2_{\gamma}\Ibb$, $\Sigma_n =\sigma^2_n \Ibb$ for some $\sigma^2_{\gamma}, \sigma^2_n>0$. We further consider action $a \in \Acal$ and context $c \in \Ccal$ to be $d=5$ dimensional vectors with $a_i$ and $c_i$ respectively denoting their $i$th component. We use $\phi(a,c)=[a_1^2, a_2^2,a_3^2,a_4^2,a_5^2,c_1^2,c_2^2,c_3^2,c_4^2,c_5^2,a_1c_1,a_2c_2,a_3c_3, a_4c_4,$ $a_5c_5]$ as the $m=15$ dimensional feature vector.\\
\textbf{Gaussian Bandits}: The mean reward function is given by $f(\thetastar,a,c)=\phi(a,c)^\top\thetastar$ with the feature map described above. Other parameters are fixed as $\sigma^2=2$ and $\lambda=0.01$. Plots are averaged over 100 independent trials.\\
{\textbf{Logistic Bandits}}: The reward $r_t \in \{0,1\}$ is Bernoulli with mean reward given by $\mu(\phi(a,c)^\top\thetastar)$, where $\mu(z) =1/(1+\exp(-z))$ is the sigmoid function. We consider a Gaussian prior $\mathcal{N}(\boldsymbol{0}, \Ibb)$ over $\thetastar$. In Algorithm~1, we choose the sampling distribution $\bar{P}_t(\thetastar) \propto P(\thetastar) \prod_{\tau=1}^{t-1}{\rm Ber}(\mu(\psi(a_{\tau},\chat_{\tau}|\Hscr_{\chat})^{\top}\thetastar))$.  However, the posterior $\bar{P}_t(\thetastar)$ is analytically intractable  since Bernoulli reward-Gaussian prior forms a non-conjugate distribution pair. Consequently, we use  Langevin Monte Carlo (LMC) \citep{xu2022langevin} to sample from $\bar{P}_t(\thetastar)$. We run LMC for $I=50$ iterations with learning rate $\eta_t=0.2/t$ and inverse temperature $\beta^{-1}=0.001$. Plots are averaged over 10 independent trials.

%The action set $\Acal$ consists of $K=40$ actions. We fix reward noise  variance to be $\sigma^2=0.01$ and the prior distribution parameters as $\lambda=0.01$ and $\Sigma_{\gamma}=0.5\Ibb$. The Gaussian noise channel $P(\chat|c,\gammastar)$ has mean $c+\gammastar$ with variance $\Sigma_n=0.5\Ibb$. 
\emph{\textbf{MovieLens Dataset}}: We use the MovieLens-100K dataset \citep{harper2015movielens} to evaluate the performances. We first perform non-negative matrix factorization on the rating matrix $R = [r_{c,a}] \in  \mathbb{R}^{943 \times 1682}$ with 3 latent factors to get $R = WH$, where $W \in \mathbb{R}^{943 \times 3}$ and  $H \in \mathbb{R}^{3\times 1682}$. Each row vector $W_c$ corresponds to an user context, while each column vector $H_a$ corresponds to movie (action) features. The mean and variance of the Gaussian context distribution is estimated from the row vectors of $W$. We then add Gaussian noise to context as in the synthetic settings with $\sigma^2_n=0.1$. 
We apply K-means algorithm to the column vector of $H$ to group the actions into $K=20$ clusters, and let $m_k$ be the centroid and $v_k$ be the variance of the $k$th cluster. The mean and variance of the Gaussian prior over $\thetastar$ are then set as $\mu_{\theta}=(m_1,\hdots,m_K)$ and $\Sigma_{\theta}={\rm diag}(v_1,\hdots,v_K)$ respectively. We consider the feature vector $\phi(a,c)$ to be  a $90$-dimensional vector with $W_c$ at the index of the cluster $k$ to which action $a$ belong. We additionally add mean-zero Gaussian noise to the true reward with variance $\sigma^2=0.01$.
%We apply the k-means algorithm to the row vectors of H to produce K = 30 groups (arms), and let θa be the center of the a-th group. Finally, we randomly choose M = 100 users’ feature vectors. We observe that 0.4 ≤ ∥xi,a∥2 ≤ 0.8, and the suboptimality gaps lie in [0.01, 0.8]. The regret performances of the algorithms are plotted in Figure 1(c). The curves show similar characteristics as in Figure 1(a). These results demonstrate the effectiveness of collaborative learning in the federated bandits setting.

\emph{\textbf{Baselines}}: We compare our algorithms with two baselines: ${\rm TS\_{\rm naive}}$ and ${\rm TS\_{\rm oracle}}$. In ${\rm TS\_{\rm naive}}$, the agent observes only noisy contexts but is unaware of the presence of noise. Consequently, it naively implements conventional TS with noisy context $\chat_t$. This sets the benchmark for the worst-case achievable regret.  The second baseline ${\rm TS\_{\rm oracle}}$ assumes that the agent knows the true channel parameter $\gammastar$, a setting studied in \citep{park2021analysis}, and can thus perform exact denoising via the predictive posterior distribution $P(c_t|\chat,\gammastar)$. 
%Consequently,  the agent uses the expected feature map $\psi(a,\chat_t|\gammastar)$ to choose the action as well as to update the posterior. 
This algorithm sets the benchmark for the best achievable regret.
%Finally, we consider the baseline  {\it TS\_kk} where the agent has knowledge of context distribution alone, as in \citep{kirschner2019stochastic}, which it uses  to evaluate the expected feature map $\Ebb_{P(c_t)}[\phi(a,c_t)]$ for action selection and posterior update. 
%We note that {\it TS\_noisycontexts} and {\it TS\_kk} correspond to worst-case regrets.

%In Fig.~\ref{fig:expt1} (Left and Center), we compare the performance of the proposed algorithms Algorithm~\ref{alg:1} and Algorithm~\ref{alg:2} with three baseline algorithms for unobserved true contexts -- {\it TS\_noisycontexts}, {\it TS\_pocmab} and {\it TS\_kk}. The baseline {\it TS\_noisycontexts} implements  TS using the observed noisy context $\chat_t$, in place of true context $c_t$, for action selection as well as posterior update, while {\it TS\_pocmab} implements the algorithm of \citep{park2021analysis} with the exact predictive distribution $P(c_t|\chat_t,\gammastar)$ known to the agent. Finally, the baseline {\it TS\_kk} neglects the observed noisy contexts and uses the knowledge of context distribution $P(c)$ to choose action and to update the posteriors as in \citep{kirschner2019stochastic}. Thus, {\it TS\_pocmab} sets the benchmark for best achievable regret, while {\it TS\_noisycontexts} and {\it TS\_kk} determine worst-case regrets.
%We run each algorithm for $100$ independent trials and plot the mean in Fig.~\ref{fig:expt1}.  Fig.~\ref{fig:expt1}(Right) compares the performance of Algorithm~\ref{alg:2} when the number $K$ of actions is varied.

Fig.~\ref{fig:expt1}(Left) corroborate our theoretical findings for Gaussian bandits. In particular, our algorithms (Alg1 and Alg2) demonstrate  sub-linear regret and achieve robust performance comparable to the best achievable performance of ${\rm TS\_{\rm oracle}}$. We remark that while our regret analysis of Gaussian bandits is motivated due to the easy tractability of posterior distributions and  the nice concentration properties of Gaussians, our empirical results for logistic bandits in Fig.~\ref{fig:expt1}(Center) show promising extension of our algorithms to non-conjugate distributions.  Extension of Bayesian regret analysis to such general distributions is left for future work.
Further, our experiments on MovieLens data in Fig.~\ref{fig:expt1}(Right) validate the effectiveness of our algorithm in comparison to the benchmarks. The plot shows that our approach outperforms ${\rm TS\_{\rm naive}}$ and achieves comparable regret as that of ${\rm TS\_{\rm oracle}}$ which is the best achievable regret.

\bibliography{Bandits,ref}
%\bibliographystyle{ieeetr}

%%%%%%%%%%%%%%%%%%%%%%%%%%%%%%%%%%%
%%%%%% SUPPLEMENT (OPTIONAL) %%%%%%
%%%%%%%%%%%%%%%%%%%%%%%%%%%%%%%%%%%

\newpage

\onecolumn

\title{Thompson Sampling for Stochastic Bandits with Noisy Contexts: An Information-Theoretic Regret Analysis\\ (Supplementary Material)}
% For two-column format, uncomment the following:
%\twocolumn[ \makesupplementtitle ]
\maketitle

\appendix

\section{Preliminaries} \label{app:preliminaries}

\begin{definition}[Sub-Gaussian Random Variable] A random variable $y$ is said to be $s^2$-sub-Gaussian with respect to the distribution $P(y)$ if the following inequality holds:
\begin{align}
\Ebb_{P(y)}[\exp(\lambda(y-\Ebb_{P(y)}[y])] \leq \exp\Bigl(\frac{\lambda^2s^2}{2}\Bigr). \label{eq:subgaussianity_definition}
\end{align}
\end{definition}
\begin{lemma}[Change of Measure Inequality]\label{lem:DVinequality}
    Let $x\in \mathbb{R}^n$ be a random vector and $g:\mathbb{R}^n \rightarrow \mathbb{R}$ denote a real-valued function. Let $P(x)$ and $Q(x)$ be two probability distributions defined on the space of $x$. If $g(x)$ is  $s^2$-sub-Gaussian with respect to $Q(x)$, then the following inequality holds,
\begin{align}
    |\Ebb_{P(x)}[ g(x)] -\Ebb_{Q(x)}[ g(x)]| \leq \sqrt{2s^2D_{\rm KL}(P(x) \Vert Q(x))}. \label{eq:changeofmeasureinequality}
\end{align}
\end{lemma}
\begin{proof}
    The inequality \eqref{eq:changeofmeasureinequality} follows by using the Donsker-Varadhan inequality \eqref{eq:DV_inequality} with $f(x)=\lambda g(x)$ for $\lambda \in \mathbb{R}$. This yields that 
\begin{align}
    D_{\rm KL}(P(x) \Vert Q(x)) &\geq \Ebb_{P(x)}[\lambda g(x)] -\log \Ebb_{Q(x)}[\exp(\lambda g(x))] \nonumber \\
    & \geq \Ebb_{P(x)}[\lambda g(x)] -\Ebb_{Q(x)}[\lambda g(x)] -\lambda^2 \frac{s^2}{2} \label{eq:com_1}
\end{align} where the last inequality follows from the assumption of sub-Gaussianity. Rearranging, we get that
\begin{align}
\Ebb_{P(x)}[\lambda g(x)] -\Ebb_{Q(x)}[\lambda g(x)] \leq \lambda^2 \frac{s^2}{2}+D_{\rm KL}(P(x) \Vert Q(x)).
\end{align}
For $\lambda>0$, we get that \begin{align}
\Ebb_{P(x)}[ g(x)] -\Ebb_{Q(x)}[ g(x)] \leq \lambda \frac{s^2}{2}+\frac{D_{\rm KL}(P(x) \Vert Q(x))}{\lambda},
\end{align} and optimizing over $\lambda>0$ then yields that
 \begin{align}
\Ebb_{P(x)}[ g(x)] -\Ebb_{Q(x)}[ g(x)] \leq \sqrt{2s^2D_{\rm KL}(P(x) \Vert Q(x))} . 
\end{align}
Similarly, for $\lambda<0$, we get that
\begin{align}
\Ebb_{Q(x)}[ g(x)] -\Ebb_{P(x)}[ g(x)] \leq \sqrt{2s^2D_{\rm KL}(P(x) \Vert Q(x))}.
\end{align}
\end{proof}
\begin{lemma} \label{lem:maxgaussian_com}
Let $x \in \mathbb{R}^n$ be distributed according to $Q(x)=\prod_{i=1}^n  \mathcal{N}(x_i|\mu_i,\sigma^2_i)$, i.e., each element of the random vector is independently distributed according to  a Gaussian distribution with mean $\mu_i$ and variance $\sigma^2_i$. Let $g(x) = \max_{i=1,\hdots n} x_i$ denote the maximum of $n$ Gaussian random variables. Then, the following inequality holds for $\lambda \geq 0$,
\begin{align}
    \log \mathbb{E}_{Q(x)}[\exp(\lambda g(x))] \leq \log n + \lambda \max_i \mu_i +\lambda^2 \frac{\max_i \sigma^2_i}{2}. \label{eq:maxgaussian}
\end{align}
For any distribution $P(x)$  that is absolutely continuous with respect to $Q(x)$, we then have the following change of measure inequality,
\begin{align}
\mathbb{E}_{P(x)}[g(x)]-\mathbb{E}_{Q(x)}[g(x)] \leq \sqrt{2 \Bigl(\log n+D_{\rm KL}(P(x) \Vert Q(x))\Bigr)\max_i \sigma^2_i}\label{eq:maxgaussian_cominequality}.
\end{align}
\end{lemma}
\begin{proof}
The proof of inequality \eqref{eq:maxgaussian} follows from standard analysis (see \citep{kim2023contextual}). We present it here for the sake of completeness. The following sequence of relations hold for any $\lambda \geq 0$,
\begin{align}
    \Ebb_{Q(x)}[\exp(\lambda g(x))] = \Ebb_{Q(x)}[\max_i \exp(\lambda x_i)] \leq \sum_{i=1}^n \Ebb_{Q(x_i)}[\exp(\lambda x_i)]&=\sum_{i=1}^n \exp\Bigl(\lambda \mu_i+\lambda^2 \sigma_i^2/2 \Bigr) \nonumber \\&\leq n \exp(\lambda \max_i \mu_i+\lambda^2 \max_i \sigma^2_i/2\Bigr).
\end{align} Taking logarithm on both sides of the inequality yields the upper bound in \eqref{eq:maxgaussian}. We now apply the DV inequality \eqref{eq:DV_inequality} as in \eqref{eq:com_1}. This yields that 
\begin{align}
    D_{\rm KL}(P(x) \Vert Q(x)) &\geq \Ebb_{P(x)}[\lambda g(x)] -\log \Ebb_{Q(x)}[\exp(\lambda g(x))] \nonumber \\
    & \stackrel{(a)}{\geq} \Ebb_{P(x)}[\lambda g(x)]-\log n - \lambda\max_i \mu_i -\lambda^2 \frac{\max_i \sigma^2_i}{2} \nonumber \\
    & \stackrel{(b)}{\geq}\Ebb_{P(x)}[\lambda g(x)]-\Ebb_{Q(x)}[\lambda g(x)]-\log n -\lambda^2 \frac{\max_i \sigma^2_i}{2} \label{eq:com_2},
\end{align} where the inequality in $(a)$ follows from \eqref{eq:maxgaussian}. The inequality in $(b)$ follows from observing that $\max_i x_i \geq x_i$ for all $i$, whereby we get that $\Ebb_{Q(x)}[\max_i x_i] \geq \mu_i$ which holds for all $i$. The latter inequality implies that $ \Ebb_{Q(x)}[\max_i x_i] \geq \max_i \mu_i$. Re-arranging and optimizing over $\lambda\geq 0$ then yields the required inequality in \eqref{eq:maxgaussian_cominequality}.
\end{proof}
\section{Linear-Gaussian Contextual Bandits with Delayed Contexts}
In this section, we provide all the details relevant to the Bayesian cumulative regret analysis of TS for delayed, linear-Gaussian contextual bandits.
\subsection{TS Algorithm for Linear-Gaussian Bandits with Delayed True Contexts}\label{app:TSalgorithm_delayed}
The pseudocode for the TS algorithm for Gaussian bandits is given in Algorithm~3.
\begin{algorithm}[h!]
\caption{\textbf{Algorithm 3}: TS with Delayed Contexts for Gaussian Bandits ($\pi^{\rm{TS}}_{\rm delay}$)}
\begin{algorithmic}[1]
\State  Given parameters: $(\Sigma_n,\sigma^2,\lambda,\Sigma_{\gamma},\mu_c,\Sigma_c)$.
 Initialize $\tilde{\mu}_0 = \boldsymbol{0} \in \mathbb{R}^m$ and $\tilde{\Sigma}^{-1}_0=(1/\lambda) \Ibb$
\For{$t=1,\hdots,T$}
\State The environment selects a true context $c_t$.
\State Agent observes noisy context $\chat_t$.
\State Agent computes $\tilde{R}_t$ and $\tilde{V}_t$ using \eqref{eq:covariance_predictedcontext_delayed} and \eqref{eq:mean_predictedcontext_delayed}  to evaluate $P(c_t|\chat_t,\Hscr_{t-1,c,\chat})=\Nscr(\tilde{V}_t,\tilde{R}_t^{-1})$.
\State Agent samples $\theta_t \sim \Nscr(\tilde{\mu}_{t-1},\tilde{\Sigma}_{t-1}^{-1})$ where $\tilde{\mu}_{t-1}$ and $\tilde{\Sigma}_{t-1}$ are defined as in \eqref{eq:mean_theta_delayed} and \eqref{eq:variance_theta_delayed}.
\State Agent chooses action $a_t$ as in \eqref{eq:action_TS_delayed} using $\theta_t$ and $P(c_t|\chat_t,\Hscr_{t-1,c,\chat})$.
\State Agent observes reward $r_t$ corresponding to $a_t$, and the true context $c_t$.
%\State 7. Update the history $\Hscr_{t,r,a,c,\chat}=\Hscr_{t-1,r,a,c,\chat} \cup (r_t,a_t,c_t,\chat_t)$
\EndFor
\end{algorithmic}
\end{algorithm}
\subsection{Derivation of Posterior and Predictive Posterior Distributions}
In this section, we provide detailed derivation of posterior predictive distribution for Gaussian bandits. To this end, we first derive the exact predictive distribution $P(c_t|\chat_t,\gammastar)$.
\subsubsection{Derivation of $P(c_t|\chat_t,\gammastar)$ }
We begin by noting that 
\begin{align}
 P(c_t|\chat_t,\gammastar) =\frac{P(c_t)P(\chat_t|c_t,\gammastar)}{P(\chat_t|\gammastar)} &\propto P(c_t)P(\chat_t|c_t,\gammastar) \nonumber \\
&= \Nscr(\mu_c,\Sigma_c) \Nscr(c_t+\gammastar,\Sigma_n). \non
\end{align} Subsequently, 
\begin{align}
   \log(P(c_t|\chat_t,\gammastar)) &\propto \log( P(c_t)P(\chat_t|c_t,\gammastar))\non\\& \propto -\frac{1}{2}\biggl((c_t-\mu_c)^\top\Sigma_c^{-1}(c_t-\mu) + (\chat_t-c_t-\gammastar)^\top \Sigma_n^{-1}(\chat_t-c_t-\gammastar)\biggr) \nonumber \\
    & = -\frac{1}{2} \biggl( c_t^\top(\Sigma_c^{-1}+\Sigma_n^{-1})c_t -\Bigl(\mu_c^\top \Sigma_c^{-1}+(\chat_t-\gammastar)^\top \Sigma_n^{-1} \Bigr)c_t \non\\
    &-c_t^\top \Bigl(\Sigma_c^{-1}\mu_c+\Sigma_n^{-1}(\chat_t-\gammastar) \Bigr)  +(\chat_t-\gammastar)^\top \Sigma_n^{-1}(\chat_t-\gammastar)\biggr) \non  \\
    & =-\frac{1}{2} \biggl( c_t^\top M c_t -A_t^\top M c_t  -c_t^\top M A +A_t^\top M A - A_t^\top M A  +(\chat_t-\gammastar)^\top \Sigma_n^{-1}(\chat_t-\gammastar)\biggr)
    \label{eq:posteriorderivation_1},
\end{align} where we have defined
\begin{align}
    M &= \Sigma_c^{-1}+\Sigma_n^{-1} \label{eq:M} \\
     A_t & =(M^{-1})^\top \Bigl( \Sigma_c^{-1}\mu_c+\Sigma_n^{-1}(\chat_t-\gammastar) \Bigr). \label{eq:A}
\end{align}
From \eqref{eq:posteriorderivation_1}, we get that 
\begin{align}
\log(P(c_t|\chat_t,\gammastar)) \propto -\frac{1}{2} \biggl( c_t^\top M c_t -A_t^\top M c_t  -c_t^\top M A +A_t^\top M A \biggr). \non
\end{align} This implies that
\begin{align}
     P(c_t|\chat_t,\gammastar) &=\Nscr(A_t,M^{-1}). \label{eq:app_c_chatgammastar}
 \end{align}
 \subsubsection{Derivation of $P(\chat_t|\gammastar)$}
 We now derive the distribution $P(\chat_t|\gammastar)$  which is defined in \eqref{eq:chat_gammastar} as
 \begin{align*}
     P(\chat_t|\gammastar)=\Ebb_{P(c_t)}[P(\chat_t|c_t,\gammastar)].
 \end{align*}Hence, $P(\chat_t|\gammastar)$ can be obtained by marginalizing the joint distribution $P(c_t)P(\chat_t|c_t,\gammastar) = P(c_t|\chat_t,\gammastar)P(\chat_t|\gammastar)$ over $c_t$.  To this end, we use \eqref{eq:posteriorderivation_1} to get,
 \begin{align*}
     \log( P(c_t)P(\chat_t|c_t,\gammastar)) = \log(P(c_t|\chat_t,\gammastar)P(\chat_t|\gammastar)) & \propto \log(P(c_t|\chat_t,\gammastar)) - \frac{1}{2} \Bigl(  - A_t^\top M A  +(\chat_t-\gammastar)^\top \Sigma_n^{-1}(\chat_t-\gammastar)\Bigr), \end{align*} which implies that
     \begin{align}
         \log (P(\chat_t|\gammastar)) & \propto - \frac{1}{2} \Bigl(  - A_t^\top M A  +(\chat_t-\gammastar)^\top \Sigma_n^{-1}(\chat_t-\gammastar)\Bigr) \non \\
         & \propto -\frac{1}{2} \biggl( \chat_t^\top \Bigl(\Sigma_n^{-1}- \Sigma_n^{-1} (M^{-1})^\top \Sigma_n^{-1}\Bigr)\chat_t -\chat_t^\top \Bigl(\Sigma_n^{-1}(M^{-1})^\top (- \Sigma_n^{-1}\gammastar+\Sigma_c^{-1}\mu_c)+\Sigma_n^{-1}\gammastar \Bigr)\nonumber \\&-\Bigl((\mu_c^\top \Sigma_c^{-1} -\gamma^{*\top}\Sigma_n^{-1})(M^{-1}) \Sigma_n^{-1}+\gamma^{*\top}\Sigma_n^{-1} \Bigr)\chat \biggr) \non \\
     & = -\frac{1}{2} \biggl( \chat_t^\top G \chat_t -F^\top G\chat_t -G^\top F \chat_t \biggr) \nonumber 
     \\ & \propto -\frac{1}{2} \biggl((\chat_t-F)^\top G(\chat_t-F)\biggr)\non, 
     \end{align} where
     \begin{align}
         G& = \Sigma_n^{-1}-\Sigma_n^{-1}(M^{-1})^\top \Sigma_n^{-1} \label{eq:G}
 \\
 F &= (G^{-1})^\top \Bigl(\Sigma_n^{-1}(M^{-1})^\top (- \Sigma_n^{-1}\gamma^{*}+\Sigma_c^{-1}\mu_c)+\Sigma_n^{-1}\gamma^{*}\Bigr)=(G^{-1})^\top (G\gamma^{*}+\Sigma_n^{-1}(M^{-1})^\top \Sigma_c^{-1}\mu_c). \label{eq:F}
     \end{align} Thus,
     \begin{align}
     P(\chat_t|\gammastar) = \Nscr(F, G^{-1}). \label{eq:app_chat_gammastar}
     \end{align}
\subsubsection{Derivation of $P(\gammastar|\Hscr_{t-1,c,\chat})$}
We now derive the posterior distribution $P(\gammastar|\Hscr_{t-1,c,\chat})$. To this end, we use Baye's theorem as 
\begin{align}
P(\gammastar|\mathcal{H}_{t-1,c,\chat}) &\propto \prod_{\tau=1}^{t-1}P(\chat_{\tau}|c_{\tau},\gammastar) P(\gammastar)\non\\
    & = \prod_{\tau=1}^{t-1} \Nscr(c_{\tau}+\gammastar,\Sigma_n) \Nscr(\boldsymbol{0},\Sigma_{\gamma}).\non
\end{align}
We then have,
\begin{align}
 \log P(\gammastar|\mathcal{H}_{t-1,c,\chat}) &\propto -\frac{1}{2} \biggl(\sum_{\tau=1}^{t-1} \biggl( (\chat_{\tau}-c_{\tau}-\gammastar)^\top \Sigma_n^{-1} (\chat_{\tau}-c_{\tau}-\gammastar)  \biggr)+ \gammastart \Sigma_{\gamma}^{-1}\gammastar \biggr)\non \\
 &=-\frac{1}{2} \biggl(\sum_{\tau=1}^{t-1} \biggl( (-\chat_{\tau}+c_{\tau}+\gammastar)^\top\Sigma_n^{-1} (-\chat_{\tau}+c_{\tau}+\gammastar)  \biggr)+ \gammastart \Sigma_{\gamma}^{-1}\gammastar \biggr)\non \\
 & \propto -\frac{1}{2} \biggl(\gammastart((t-1)\Sigma_n^{-1}+\Sigma_{\gamma}^{-1})\gammastar-\gammastart \Sigma_n^{-1}(\sum_{\tau=1}^{t-1}(\chat_{\tau}-c_{\tau})) - (\sum_{\tau=1}^{t-1}(\chat_{\tau}-c_{\tau}))^\top \Sigma_n^{-1}\gammastar \biggr).\non
\end{align}
Consequently, we get that,
\begin{align}
P(\gammastar|\Hscr_{t-1,c,\chat})= \Nscr(\gammastar|Y_t,W_t^{-1}) \label{eq:app_posterior_delayed}
\end{align}
where 
\begin{align}
    W_t &= (t-1)\Sigma_n^{-1}+\Sigma_{\gamma}^{-1} \label{eq:W}\\
    Y_t &= (W_t^{-1})^\top\Sigma_n^{-1}\sum_{\tau=1}^{t-1}(\chat_{\tau}-c_{\tau}). \label{eq:Y}
\end{align}
\subsubsection{Derivation of Posterior Predictive Distribution $P(c_t|\chat_t,\Hscr_{t-1,c,\chat})$}\label{app:derivations_delayed}
Using results from previous subsections, we are now ready to derive the posterior predictive distribution $P(c_t|\chat_t,\Hscr_{t-1,c,\chat})$. Note that $P(c_t|\chat_t,\Hscr_{t-1,c,\chat})=\Ebb_{P(\gammastar|\Hscr_{t-1,c,\chat})}[P(c_t|\chat_t,\gammastar)]$. We then have the following set of relations:
\begin{align}
   & \log \Bigl( P(\gammastar|\Hscr_{t-1,c,\chat}) P(c_t|\chat_t,\gammastar)\Bigr) \nonumber \\&\propto -\frac{1}{2} \Bigl( (c_t-A_t)^\top M (c_t-A_t)+ (\gammastar -Y_t)^\top W_t(\gammastar-Y_t) \Bigr) \nonumber \\
    & = -\frac{1}{2} \Bigl( (c_t-D-E_t+ (M^{-1})^\top \Sigma_n^{-1}\gammastar)^\top M (c_t-D-E_t+ (M^{-1})^\top \Sigma_n^{-1}\gammastar)+ (\gammastar -Y_t)^\top W_t(\gammastar-Y_t) \Bigr) \nonumber \\
    & \propto -\frac{1}{2} \Bigl( \gammastart (\Sigma_n^{-1}(M^{-1})^\top \Sigma_n^{-1}+W_t)\gammastar - \gammastart (\Sigma_n^{-1}(D+E_t-c_t)+W_tY_t) - ( (D+E_t-c_t)^\top \Sigma_n^{-1}+Y_t^\top W_t)\gammastar \nonumber \\& + (c_t-D-E_t)^\top M (c_t-D-E_t)\Bigr) \nonumber \\
    & = -\frac{1}{2} \Bigl( \gammastart \tilde{H}_t \gammastar - \gammastart \tilde{H}_t^\top \tilde{J}_t - \tilde{J}_t^\top \tilde{H}_t \gammastar + \tilde{J}_t^\top \tilde{H}_t \tilde{J}_t - \tilde{J}_t^\top \tilde{H}_t \tilde{J}_t+ (c_t-D-E_t)^\top M (c_t-D-E_t)\Bigr) \nonumber \\
    & \propto \log(P(\gammastar|\Hscr_{t,c,\chat})) -\frac{1}{2} \Bigl(- \tilde{J}_t^\top \tilde{H}_t \tilde{J}_t+ (c_t-D-E_t)^\top M (c_t-D-E_t)\Bigr) \nonumber 
\end{align} where $D= (M^{-1})^\top \Sigma_c^{-1}\mu_c$, $E_t=(M^{-1})^\top \Sigma_n^{-1}\chat_t$, $\tilde{H}_t=\Sigma_n^{-1}(M^{-1})^\top \Sigma_n^{-1}+W_t$ and $\tilde{J}_t =(\tilde{H}_t^{-1})^\top (\Sigma_n^{-1}(D+E_t-c_t)+W_tY_t)$. 

Since 
$\log \Bigl( P(\gammastar|\Hscr_{t-1,c,\chat}) P(c_t|\chat_t,\gammastar)\Bigr)= \log \Bigl(P(c_t|\chat_t,\Hscr_{t-1,c,\chat})P(\gammastar|\Hscr_{t,c,\chat}) \Bigr)$, we get that
\begin{align}
    \log(P(c_t|\chat_t,\Hscr_{t-1,c,\chat})) &\propto -\frac{1}{2} \Bigl(- \tilde{J}_t^\top \tilde{H}_t \tilde{J}_t+ (c_t-D-E_t)^\top M (c_t-D-E_t)\Bigr) \nonumber \\
    & \propto -\frac{1}{2} \Bigl(c_t^\top \Bigl(M-\Sigma_n^{-1} (\tilde{H}_t^{-1})^\top \Sigma_n^{-1} \Bigr)c_t - c_t^\top \Bigl(M(D+E_t)-\Sigma_n^{-1}(\tilde{H}_t^{-1})^\top (\Sigma_n^{-1}(D+E_t)+W_tY_t)\Bigr) \nonumber \\
    & -\Bigl(M(D+E_t)-\Sigma_n^{-1}(\tilde{H}_t^{-1})^\top (\Sigma_n^{-1}(D+E_t)+W_tY_t) \Bigr)^\top c_t \Bigr). \nonumber
\end{align} This gives that
\begin{align}
    P(c_t|\chat_t,\Hscr_{t-1,c,\chat}) & = \Nscr(\tilde{V}_t, \tilde{R}_t^{-1})  \label{eq:app_posterior_predictive_delayed}\quad \mbox{where}\\
    \tilde{R}_t & = M-\Sigma_n^{-1} (\tilde{H}_t^{-1})^\top \Sigma_n^{-1} \label{eq:tildeR}\\
    \tilde{V}_t & = ( \tilde{R}_t^{-1})^\top \Bigl(M(D+E_t)-\Sigma_n^{-1}(\tilde{H}_t^{-1})^\top (\Sigma_n^{-1}(D+E_t)+W_tY_t)\Bigr). \label{eq:tildeV}
\end{align}
\subsection{Proof of Lemma~\ref{lem:R_CB_delayed}}
We now present the proof of Lemma~\ref{lem:R_CB_delayed}. To this end, we first recall that  $\Ebb_t[\cdot]=\Ebb[\cdot|\Fcal_t]$ where $\Fcal_{t}=\Hscr_{t-1,r,a,c,\chat}\cup \chat_t$, and we denote $P_t(\thetastar)$ as the posterior distribution $P(\thetastar|\Hscr_{t-1,r,a,c})$ of $\thetastar$ given the history of observed reward-action-context tuples.  We can then equivalently write $\Rcal^T_{\rm d,CB}$ as
\begin{align}
  \Rcal^T_{\rm d,CB}&=  \sum_{t=1}^T\mathbb{E}\Bigl[\mathbb{E}_t\Bigl[\psi(\hat{a}_t,\chat_t|\Hscr_{c,\chat})^\top\thetastar-\psi(a_t,\chat_t|\Hscr_{c,\chat})^\top\thetastar \Bigr]  \Bigr] \nonumber \\
  &=\sum_{t=1}^T \Ebb \biggl[\Ebb_t\underbrace{\Bigl[f(\thetastar,\hat{a}_t,c_t) -f(\thetastar,a_t,c_t) \Bigr]}_{:=\Delta_t} \biggr],
\end{align}where $\psi(a,\chat_t|\Hscr_{c,\chat})=\Ebb_{P(c_t|\chat_t,\Hscr_{t-1,c,\chat})}[\phi(a,c_t)]$ is as defined in \eqref{eq:action_TS_delayed} and  we have used $f(\thetastar,a_t,c_t)=\phi(a_t,c_t)^\top\thetastar$ to denote the mean-reward function.
To obtain an upper bound on $\Rcal^T_{\rm d,CB}$, we define the following {\it lifted information ratio} as in \citep{neu2022lifting}, 
\begin{align}
    \Gamma_t = \frac{(\Ebb_t[\Delta_t])^2}{\Lambda_t} \label{eq:liftedinformatinratio}
\end{align}where
\begin{align}
\Lambda_t=\Ebb_t\Bigl[\Bigl(f(\thetastar,a_t,c_t)-\bar{f}(a_t,c_t)\Bigr)^2\Bigr],
\end{align}with  $\bar{f}(a_t,c_t)=\Ebb_t[f(\thetastar,a_t,c_t)|a_t,c_t]$ denoting the expectation of mean reward with respect to the posterior distribution $P_t(\thetastar)$.
Subsequently, we get the following upper bound 
\begin{align}
    \Rcal^T_{\rm d,CB} \leq \sum_{t=1}^T \Ebb \biggl[\sqrt{\Gamma_t \Lambda_t} \biggr] \leq \sqrt{\Ebb\Bigl[\sum_{t=1}^T \Gamma_t \Bigr] \Bigl[\sum_{t=1}^T \Ebb \Bigl[\Lambda_t\Bigr] \Bigr]}, \label{eq:templateub}
\end{align}where the last inequality follows by an application of Cauchy-Schwarz inequality. An upper bound on $\Rcal^T_{\rm d,CB}$ then follows by obtaining an upper bound on the lifted information ratio $\Gamma_t$ as well as on $\Lambda_t.$

We first evaluate the term $\Lambda_t$. To this end, note that $\bar{f}(a_t,c_t)=\phi(a_t,c_t)^\top \tilde{\mu}_{t-1}$, with $\tilde{\mu}_{t-1}$ defined as in \eqref{eq:mean_theta_delayed}. Using this, we get
\begin{align}
 \Lambda_t &= \Ebb_t\Bigl[ (\phi(a_t,c_t)^\top(\thetastar-\tilde{\mu}_{t-1}))^2\Bigr]  \nonumber \\
 & = \Ebb_t \Bigl[ \phi(a_t,c_t)^\top (\thetastar -\tilde{\mu}_{t-1}) (\thetastar-\tilde{\mu}_{t-1})^\top \phi(a_t,c_t)\Bigr] \\
 & = \Ebb_t \Bigl[ \phi(a_t,c_t)^\top \tilde{\Sigma}_{t-1}^{-1} \phi(a_t,c_t)\Bigr] = \Ebb_t \Bigl[ \Vert \phi(a_t,c_t)\Vert_{ \tilde{\Sigma}_{t-1}^{-1}}^2\Bigr]
\end{align}where $\tilde{\Sigma}_{t-1}$ is as in \eqref{eq:variance_theta_delayed}, and  the third equality follows since conditional on $\mathcal{F}_t$, $(a_t,c_t)$ is independent of $\thetastar$. Subsequently, we can apply the elliptical potential lemma \citep[Lemma 19.4]{lattimore2020bandit} using the assumption that $\Vert \phi(\cdot,\cdot)\Vert_2 \leq 1$ and that $\sigma^2 /\lambda \geq 1$.  This results in
\begin{align}
    \sum_{t=1}^T \Vert \phi(a_t,c_t) \Vert _{\tilde{\Sigma}_{t-1}^{-1}}^2 &=\sigma^2  \sum_{t=1}^T \Vert \phi(a_t,c_t) \Vert _{(\sigma^2\tilde{\Sigma}_{t-1})^{-1}}^2 \nonumber \\& \leq 2\sigma^2 \log \frac{{\rm det}(\tilde{\Sigma}_T)}{{\rm det}(\sigma^2/\lambda\Ibb)} = 2\sigma^2\biggl( m \log \Bigl(\sigma^2/\lambda+ T/m\Bigr)-m \log(\sigma^2/\lambda) \biggr)=2m\sigma^2 \log \Bigl(1+\frac{T\lambda}{m\sigma^2}\Bigr). \label{eq:Lambda_ub}
\end{align}

To upper bound the lifted information ratio term $\Gamma_t$, we can use \citep[Lemma 7]{neu2022lifting}. To demonstrate how to leverage results from \citep{neu2022lifting}, we start by showing that the inequality $\Gamma_t\leq m$ holds. To this end, we note that the lifted information ratio can be equivalently written as
\begin{align}
    \Gamma_t = \frac{\biggl(\Ebb_t\Bigl[f(\theta_t,a_t,c_t)-\bar{f}(a_t,c_t)\Bigr] \biggr)^2}{\Lambda_t}
\end{align} which follows since
\begin{align}
    \Ebb_t[\Delta_t]&= \Ebb_t\Bigl[f(\thetastar,\hat{a}_t,c_t)-f(\thetastar,a_t,c_t)\Bigr] \nonumber \\
    &=\Ebb_t\Bigl[f(\thetastar,\hat{a}_t,c_t)-\bar{f}(a_t,c_t)\Bigr]\nonumber \\
    &= \sum_{a'}P_t(\hat{a}_t=a')\Ebb_t\Bigl[f(\thetastar,a',c_t)|\hat{a}_t=a'\Bigr]-\sum_{a'}P_t(a_t=a')\Ebb_t\Bigl[\bar{f}(a_t=a',c_t)\Bigr] \nonumber \\
    & = \sum_{a'}P_t(\hat{a}_t=a')\Bigl(\Ebb_t\Bigl[f(\thetastar,a',c_t)|\hat{a}_t=a'\Bigr]-\Ebb_t[\bar{f}(a',c_t)]\Bigr),
\end{align} where the second equality holds since conditioned on $\mathcal{F}_t$, $(a_t,c_t)$ and $\thetastar$ are independent. In the third equality, we denote $P_t(\hat{a}_t)=P(\hat{a}_t|\mathcal{F}_t)$ and $P_t(a_t)=P(a_t|\mathcal{F}_t)$. Using these, the last equality follows since $a_t \stackrel{d}{=}\hat{a}_t$, i.e, $P_t(a_t)=P_t(\hat{a}_t)$.
Now, let us define a $K \times K$ matrix $M$ with entries given by
\begin{align}
    M_{a,a'}= \sqrt{P_t(\hat{a}_t=a')P_t(a_t=a)}\Bigl(\Ebb_t\Bigl[f(\thetastar,a,c_t)|\hat{a}_t=a'\Bigr]-\Ebb_t[\bar{f}(a,c_t)]\Bigr).
\end{align}Using this and noting that $P_t(\hat{a}_t)=P_t(a_t)$, we get that $\Ebb_t[\Delta_t]={\rm Tr}(M)$. We now try to bound $\Lambda_t$ in terms of the matrix $M$. To see this, we can equivalently write $\Lambda_t$ as
\begin{align}
    \Lambda_t&=\sum_a P_t(a_t=a) \Ebb_t\Bigl[\Bigl(f(\thetastar,a,c_t)-\bar{f}(a,c_t) \Bigr)^2\Bigr]\nonumber \\
    & = \sum_{a,a'} P_t(a_t=a) P_t(\hat{a}_t=a')\Ebb_t\Bigl[\Bigl(f(\thetastar,a,c_t)-\bar{f}(a,c_t) \Bigr)^2|\hat{a}_t=a'\Bigr] \nonumber \\
    & \geq \sum_{a,a'} P_t(a_t=a) P_t(\hat{a}_t=a')\Bigl(\Ebb_t\Bigl[f(\thetastar,a,c_t)-\bar{f}(a,c_t) |\hat{a}_t=a'\Bigr] \Bigr)^2\nonumber\\
    &= \sum_{a,a'} P_t(a_t=a) P_t(\hat{a}_t=a')\Bigl(\Ebb_t[f(\thetastar,a,c_t)|\hat{a}_t=a']-\Ebb_t[\bar{f}(a,c_t)]\Bigr)^2=\Vert M \Vert_F^2,
\end{align}where the inequality follows by the application of Jensen's inequality. We thus get that 
$$ \Gamma_t \leq \frac{{\rm Tr}(M)^2}{\Vert M \Vert_F^2} \leq m, $$ where the last inequality follows from \citep[Prop. 5]{russo2016information}. From \citep[Lemma 3]{neu2022lifting}, we also get that $\Gamma_t \leq 2\log(1+K)$. This results in the upper bound
$\Gamma_t \leq \min\{m,2\log(1+K)\}$.

Using this and the bound of \eqref{eq:Lambda_ub} in \eqref{eq:templateub}, we get that
\begin{align}
\Rcal^T_{\rm d,CB} \leq \sqrt{2Tm\sigma^2\min\{m, 2(1+\log K)\}\log \Bigl(1+\frac{T\lambda}{m\sigma^2}\Bigr)}.
\end{align}

\subsection{Proof of Lemma~\ref{lem:EE_1}}
We now prove an upper bound on the term $\Rcal^T_{{\rm d,EE1}}$. To this end, let us define the following event: 
\begin{align}
    \Escr :=\biggl \lbrace \Vert \thetastar \Vert_2 \leq \sqrt{2\lambda m \log \Bigl(\frac{2m}{\delta} \Bigr)}:=U \biggr \rbrace. \label{eq:event}
\end{align}Note that since $\thetastar \sim \Nscr(\thetastar| \boldsymbol{0},\lambda \Ibb)$, we get that with probability at least $1-\delta$, the following inequality holds $\Vert \thetastar \Vert_{\infty} \leq \sqrt{2\lambda \log \Bigl(\frac{2m}{\delta} \Bigr)}$. Since $\Vert \thetastar \Vert_{2} \leq \sqrt{m}\Vert \thetastar \Vert_{\infty}$, the above inequality in turn implies the event $\Escr$ such that $P(\Escr)\geq 1-\delta$. 
\begin{align}
  \Rcal^T_{{\rm d,EE1}} &\stackrel{(a)}\leq   \sum_{t=1}^T \Ebb \Bigl[ \psi(a^{*}_t,\chat_t|\gammastar)^\top \thetastar -\psi(a^*_t,\chat_t|\Hscr_{c,\chat})^\top\thetastar \Bigr] \nonumber \\
  &= \sum_{t=1}^T \Ebb \Bigl[ \underbrace{\Ebb_{P(c_t|\chat_t,\gammastar)}[\phi(a^{*}_t,c_t)^\top \thetastar] -\Ebb_{P(c_t|\chat_t,\Hscr_{c,\chat})}[\phi(a^{*}_t,c_t)^\top \thetastar]}_{:=\Delta\Bigl(P(c_t|\chat_t,\gammastar), P(c_t|\chat_t,\Hscr_{c,\chat})\Bigr)} \Bigr],\nonumber \\
  & = \sum_{t=1}^T \Ebb \Bigl[ \Delta\Bigl(P(c_t|\chat_t,\gammastar), P(c_t|\chat_t,\Hscr_{c,\chat})\Bigr) \indicator\{\Escr\}] + \sum_{t=1}^T \Ebb \Bigl[ \Delta\Bigl(P(c_t|\chat_t,\gammastar), P(c_t|\chat_t,\Hscr_{c,\chat})\Bigr) \indicator\{\Escr^c\}] \\
  & \stackrel{(b)}{\leq} \sum_{t=1}^T \Ebb \Bigl[ \Delta\Bigl(P(c_t|\chat_t,\gammastar), P(c_t|\chat_t,\Hscr_{c,\chat})\Bigr) \indicator\{\Escr\}] + 2\delta T \Ebb[\Vert \thetastar \Vert_2 |\Escr^c] \label{eq:EE1_delayed_1},
\end{align}where the inequality $(a)$ follows from the definition of $\hat{a}_t =\arg \max_{a \in \Acal} \psi(a,\chat|\Hcal_{c,\chat})^\top \thetastar$, and $\indicator\{\bullet\}$ denotes the indicator function which takes value $1$ when $\bullet$ is true and takes value $0$ otherwise. The inequality in $(b)$ follows by noting that
\begin{align*}
   \Ebb[\Delta\Bigl(P(c_t|\chat_t,\gammastar), P(c_t|\chat_t,\Hscr_{c,\chat})\Bigr) \Ibb\{\Escr^c\}] &\leq \Ebb\Bigl[ \Bigl(\Ebb_{P(c_t|\chat_t,\gammastar)}[\phi(a^*_t,c_t)]-\Ebb_{P(c_t|\chat_t,\Hscr_{c,\chat})}[\phi(a^*_t,c_t)]\Bigr)^\top \thetastar \indicator\{\Escr^c\} \Bigr] \non \\
   & \leq \Ebb\Bigl[ \Vert \Ebb_{P(c_t|\chat_t,\gammastar)}[\phi(a^*_t,c_t)]-\Ebb_{P(c_t|\chat_t,\Hscr_{c,\chat})}[\phi(a^*_t,c_t)]\Vert_2 \Vert \thetastar \Vert_2\indicator\{\Escr^c\} \Bigr] \non \\
   & \leq 2\Ebb\Bigl[\Vert \thetastar \Vert_2\indicator\{\Escr^c\} \Bigr] = 2 P(\Escr^c) \Ebb[\Vert \thetastar \Vert_2 |\Escr^c] \leq 2 \delta \Ebb[\Vert \thetastar \Vert_2 |\Escr^c],
\end{align*}where the last inequality is due to $P(\Escr^c)=1-P(\Escr) \leq \delta$. To obtain an upper bound on $\Ebb[\Vert \thetastar \Vert_2 |\Escr^c]$, we note that the following set of inequalities hold:
\begin{align}
    \Ebb[\Vert \thetastar \Vert_2 |\Escr^c] &\stackrel{(a)}{\leq} \sqrt{m} \Ebb[\Vert \thetastar \Vert_{\infty} |\Escr^c] \stackrel{(b)}{=} \sqrt{m} \Ebb[\Vert \thetastar \Vert_{\infty} |\Vert \thetastar \Vert_{\infty}>u] \nonumber \\
    & = \sqrt{m} \sum_{i=1}^m P(\Vert \thetastar \Vert_{\infty}=|\thetastar_i|) \Ebb\Bigl[\Vert \thetastar \Vert_{\infty} \Bigl|\Vert \thetastar \Vert_{\infty}=|\thetastar_i|, \Vert \thetastar \Vert_{\infty}>u \Bigr] \non \\
    & \leq \sqrt{m} \sum_{i=1}^m \Ebb\Bigl[|\thetastar_i| \Bigl| |\thetastar_i|>u\Bigr] \stackrel{(c)}{=}2\sqrt{m} \sum_{i=1}^m \int_{x>u} x g(x) dx \nonumber \\
    & \stackrel{(d)}{=} -2\lambda\sqrt{m} \sum_{i=1}^m \int_{x>u}  g'(x) dx = 2\lambda m^{3/2}g(u) = 2\lambda m^{3/2}\frac{1}{\sqrt{2\pi \lambda}}\exp(-u^2/2\lambda) \non \\
    & = \delta \sqrt{\frac{m\lambda}{2\pi}},
\end{align}where $(a)$ follows since $\Vert \theta \Vert_2 \leq \sqrt{m}\Vert \theta \Vert_{\infty}$, $(b)$ follows since $ \Vert \thetastar \Vert_{2} > \sqrt{m}\sqrt{2\lambda \log \Bigl(\frac{2m}{\delta}\Bigr)}$ implies that $\Vert \thetastar \Vert_{\infty} > \sqrt{2\lambda \log \Bigl(\frac{2m}{\delta}\Bigr)}:=u $. The equality in $(c)$ follows by noting that $|\thetastar_i|$, where $\thetastar_i \sim \Nscr(0,\lambda)$, follows a folded Gaussian distribution with density $2g(\thetastar_i)$ where $g(\thetastar_i)=\frac{1}{\sqrt{2\pi \lambda}}\exp(-\theta_i^{*2}/(2\lambda))$ is the Gaussian density. The equality in $(d)$ follows by noting that $xg(x)=-\lambda g'(x)$, where $g'(x)$ is the derivative of the Gaussian density. Thus, we have the following upper bound
\begin{align}
    \sum_{t=1}^T \Ebb[\Delta\Bigl(P(c_t|\chat_t,\gammastar), P(c_t|\chat_t,\Hscr_{c,\chat})\Bigr) \indicator\{\Escr^c\}] &\leq 2T \delta^2 \sqrt{\frac{m\lambda}{2\pi}}.
\end{align}

We now obtain an upper bound on $ \sum_{t=1}^T \Ebb \Bigl[ \Delta\Bigl(P(c_t|\chat_t,\gammastar), P(c_t|\chat_t,\Hscr_{c,\chat})\Bigr) \indicator\{\Escr\}\Bigr]$. To this end, note that
\begin{align}
     \Ebb \Bigl[ \Delta\Bigl(P(c_t|\chat_t,\gammastar), P(c_t|\chat_t,\Hscr_{c,\chat})\Bigr) \indicator\{\Escr\}\Bigr] &\leq  P(\Escr) \Ebb \Bigl[ \Delta\Bigl(P(c_t|\chat_t,\gammastar), P(c_t|\chat_t,\Hscr_{c,\chat})\Bigr) | \Escr\Bigr] \non \\
     & \leq \Ebb \Bigl[ \Delta\Bigl(P(c_t|\chat_t,\gammastar), P(c_t|\chat_t,\Hscr_{c,\chat})\Bigr) | \Escr].
\end{align}
Note that under the event $\Escr$, we have the following relation, $|\phi(a^*_t,c_t)^\top\thetastar)| \leq \Vert \phi(a^*_t,c_t)\Vert_2 \Vert \thetastar \Vert_2 \leq U$, whereby $\phi(a^*_t,c_t)^\top\thetastar$ is $U^2$-sub-Gaussian.
\begin{comment}Note that the prior distribution on  $\thetastar$ is Gaussian $\Nscr(\thetastar|\boldsymbol{0},\lambda \Ibb)$. Consequently,  using \citep[Theorem 1]{hsu2012tail}, we get that with probability at least $1-\delta$, the following inequality holds,
\begin{align}
    \Vert \thetastar \Vert^2_2 \leq \lambda \Bigl(m+2\sqrt{m\log \frac{1}{\delta}}+2 \log \frac{1}{\delta}\Bigr):=U.
\end{align}
This implies that with probability at least $1-\delta$, we have the inequality $|\phi(a^*_t,c_t)^\top\thetastar)| \leq \Vert \phi(a^*_t,c_t)\Vert \Vert \thetastar \Vert \leq \sqrt{U}$, whereby $\phi(a^*_t,c_t)^\top\thetastar$ is $U$-sub-Gaussian. Note that in the above inequality we used  that the feature map has bounded norm.
\end{comment}
Consequently, applying Lemma~\ref{lem:DVinequality} gives the following upper bound
\begin{align}
   \sum_{t=1}^T  \Ebb \Bigl[ \Delta\Bigl(P(c_t|\chat_t,\gammastar), P(c_t|\chat_t,\Hscr_{c,\chat})\Bigr) \indicator\{\Escr\}] &\leq \sum_{t=1}^T \Ebb\Bigl[ \sqrt{2U^2 D_{\rm KL}(P(c_t|\chat_t,\gammastar)\Vert P(c_t|\chat_t,\Hcal_{c,\chat})}\Bigr] \nonumber \\
   & \leq \sqrt{2TU^2 \sum_{t=1}^T \Ebb\Bigl[D_{\rm KL}(P(c_t|\chat_t,\gammastar)\Vert P(c_t|\chat_t,\Hcal_{c,\chat})\Bigr]}\nonumber \\
   & \stackrel{(a)}{=}\sqrt{2TU^2 \sum_{t=1}^T I(c_t;\gammastar|\chat_t,\Hcal_{c,\chat})}\nonumber\\
   & \stackrel{(b)}{\leq} \sqrt{2TU^2 \sum_{t=1}^T I(c_t,\chat_t;\gammastar|\Hcal_{c,\chat})}\\
   & \stackrel{(c)}{=} \sqrt{2TU^2  I(\Hcal_{T,c,\chat};\gammastar)} \label{eq:EE1_ub_app}
\end{align} where the equality in $(a)$ follows by the definition of condition mutual information $$I(c_t;\gammastar|\chat_t,\Hcal_{c,\chat}):=\Ebb\Bigl[D_{\rm KL}\Bigl(P(c_t,\gammastar|\chat_t,\Hcal_{c,\chat}) \Vert P(c_t|\chat_t,\Hcal_{c,\chat})P(\gammastar|\chat_t,\Hcal_{c,\chat})\Bigr)\Bigr],$$ and inequality in $(b)$ follows since $I(c_t,\chat_t;\gammastar|\Hcal_{c,\chat})=I(\chat_t;\gammastar|\Hcal_{c,\chat})+I(c_t;\gammastar|\chat_t,\Hcal_{c,\chat}) \geq I(c_t;\gammastar|\chat_t,\Hcal_{c,\chat})$ due to the non-negativity of mutual information, and finally, the equality in $(c)$ follows from the chain rule of mutual information.

We now analyze the mutual information $I(\gammastar; \Hcal_{T,c,\chat})$ which can be written as 
\begin{align}
I(\gammastar; \Hcal_{T,c,\chat})&=H(\gammastar)-H(\gammastar|\Hcal_{T,c,\chat}) \nonumber \\
& =\frac{1}{2} \log\Bigl( (2\pi e)^d {\rm det}(\Sigma_{\gamma})\Bigr) - \frac{1}{2} \log\Bigl( (2\pi e)^d {\rm det}(W^{-1})\Bigr)\\
&=\frac{1}{2} \log \frac{{\rm det}(\Sigma_{\gamma})}{{\rm det}(W^{-1})}
\end{align} where $W =(T-1)\Sigma_n^{-1}+\Sigma_{\gamma}^{-1} =\Sigma_{\gamma}^{-1}(\Ibb + (T-1)\Sigma_{\gamma}\Sigma_{n}^{-1})$. Using this we can equivalently write
\begin{align}
I(\gammastar; \Hcal_{T,c,\chat})&=\frac{1}{2} \log \frac{1}{{\rm det}((\Ibb + (T-1)\Sigma_{\gamma}\Sigma_{n}^{-1})^{-1})} \\
& = \frac{1}{2} \log {\rm det}(\Ibb + (T-1)\Sigma_{\gamma}\Sigma_{n}^{-1}).
\end{align}
If $\Sigma_{\gamma}=\sigma_{\gamma}^2 \Ibb$ and $\Sigma_{n}=\sigma_{n}^2 \Ibb$, we get that
\begin{align}
I(\gammastar; \Hcal_{T,c,\chat})&=\frac{1}{2} d \log \Bigl( 1+(T-1)\frac{\sigma^2_{\gamma}}{\sigma^2_n}\Bigr).
\end{align}
Using this in \eqref{eq:EE1_ub_app}, we get that 
\begin{align}
\sum_{t=1}^T  \Ebb \Bigl[ \Delta\Bigl(P(c_t|\chat_t,\gammastar), P(c_t|\chat_t,\Hscr_{c,\chat})\Bigr) \Ibb\{\Escr\}] \leq \sqrt{Td\log \Bigl( 1+(T-1)\frac{\sigma^2_{\gamma}}{\sigma^2_n}\Bigr) U^2}.
\end{align}
Finally, using this in \eqref{eq:EE1_delayed_1}, gives the following upper bound
\begin{align}
    \Rcal^T_{{\rm d,EE1}} \leq \sqrt{2\lambda m Td\log \Bigl( 1+(T-1)\frac{\sigma^2_{\gamma}}{\sigma^2_n}\Bigr)  \log \Bigl(\frac{2m}{\delta} \Bigr) }+2T \delta^2 \sqrt{\frac{m\lambda}{2\pi}}.
\end{align}

We finally note that same upper bound holds for the term $\Rcal^T_{\rm d,EE2}$.
\section{Linear-Gaussian Noisy Contextual Bandits with Unobserved True Contexts}
For notational simplicity, throughout this section, we use $\psi(a):=\psi(a,\chat_t|\Hscr_{\chat})=\Ebb_{P(c_t|\chat_t,\Hscr_{t-1,\chat})}[\phi(a,c_t)]$ to denote the expected feature map. Furthermore, we use $\mathcal{F}_t =\Hscr_{t-1,r,a,\chat} \cup \chat_t$.
\subsection{Derivation of Posterior Predictive Distribution}
In this section, we derive the posterior predictive distribution $P(c_t|\chat_t,\Hscr_{t-1,\chat})$ for Gaussian bandits with Gaussian context noise. To this end, we first derive the posterior $P(\gammastar|\Hscr_{t-1,\chat})$.
\subsubsection{Derivation of posterior $P(\gammastar|\Hscr_{t-1,\chat})$ }
Using Baye's theorem, we have
\begin{align}
    P(\gammastar|\Hscr_{t-1,\chat}) \propto P(\gammastar) \prod_{\tau=1}^{t-1}P(\chat_{\tau}|\gammastar), \non
\end{align} where $P(\chat_{\tau}|\gammastar)$ is derived in \eqref{eq:app_chat_gammastar}. Subsequently, we have that
 \begin{align}
 \log p(\gamma^{*}|\Hscr_{t-1},\chat) &\propto -\frac{1}{2}\sum_{\tau=1}^{t-1}\biggl( (\chat_{\tau}-F)^\top G(\chat_{\tau}-F) \biggr)-\frac{1}{2}\Bigl(\gammastart \Sigma_{\gamma}^{-1}\gammastar \Bigr) \non\\
& \propto -\frac{1}{2}\biggl(\gammastart((t-1)G+\Sigma_{\gamma}^{-1})\gammastar-\gammastart(G\chat_{1:t-1}-(t-1)\Sigma_n^{-1}(M^{-1})^\top \Sigma_c^{-1}\mu_c)\nonumber \\& -\Bigl(\chat_{1:t-1}^\top G-(t-1)\mu_c^\top \Sigma_c^{-1}M^{-1}\Sigma_n^{-1} \Bigr)\gammastar \biggr)\non,
 \end{align}where we have denoted $\sum_{\tau=1}^{t-1}\chat_{\tau}=\chat_{1:t-1}$. We then get that
 \begin{align}
 P(\gammastar|\Hscr_{t-1,\chat}) &= \Nscr(\tilde{M}_t, N_t^{-1} ) \quad \mbox{where}, \label{eq:app_posterior_nodelay}\\
 N_t &=(t-1)G+\Sigma_{\gamma}^{-1} \label{eq:N}\\
 \tilde{M}_t & = (N_t^{-1})^\top \Bigl(G\chat_{1:t-1}-(t-1)\Sigma_n^{-1}(M^{-1})^\top \Sigma_c^{-1}\mu_c \Bigr)\label{eq:tildeM}.
 \end{align}
 \subsubsection{Derivation of $P(c_t|\chat_t,\Hscr_{t-1,\chat})$}\label{app:posterior_predictive_nodelay}
 The derivation of posterior predictive distribution $P(c_t|\chat_t,\Hscr_{t-1,\chat})$ follows in a similar line as that in Appendix~\ref{app:derivations_delayed}. We start the derivation by noting that $P(c_t|\chat_t, \Hscr_{t-1,\chat}) = \Ebb_{P(\gamma^{*}|\Hscr_{t-1,\chat})}[P(c_t|\chat_t,\gamma^{*})]$.

We have that
\begin{align}
&\log( P(\gamma^{*}|\Hscr_{t-1,\chat})P(c_t|\chat_t,\gamma^{*}))\nonumber \\ &\propto -\frac{1}{2} \biggl((c_t-A_t)^\top M(c_t-A_t)+ (\gammastar-\tilde{M}_t)^\top N(\gammastar-\tilde{M}_t) \biggr) \nonumber  \\
%&= -\frac{1}{2} \biggl(\Bigl(c_t-D-(M^{-1})^\top \Sigma_n^{-1}(\chat_t-\gammastar)\Bigr)^\top M\Bigl(c_t-D-(M^{-1})^\top \Sigma_n^{-1}(\chat_t-\gammastar)\Bigr)\nonumber \\&+ (\gammastar-\tilde{M}_t)^\top N(\gammastar-\tilde{M}_t) \biggr)\\
&= -\frac{1}{2} \biggl(\Bigl(M^{-1})^\top \Sigma_n^{-1} \gammastar+ c_t-D-E_t\Bigr)^\top M\Bigl((M^{-1})^\top \Sigma_n^{-1} \gammastar+ c_t-D-E_t\Bigr)+ (\gammastar-\tilde{M}_t)^\top N_t (\gammastar-\tilde{M}_t) \biggr) \nonumber \\
%& =  -\frac{1}{2} \biggl(\gammastar^T\Sigma_n^{-1}(M^{-1})^T\Sigma_n^{-1} \gamma^{*}+\gammastar^T\Sigma_n^{-1}(c_t-D-E)+ (c_t-D-E)^T\Sigma_n^{-1}\gammastar \nonumber \\& + (c_t-D-E)^TM(c_t-D-E)+\gammastar^TN\gammastar -\tilde{M}^TN\gammastar-\gammastar^TN\tilde{M}+\tilde{M}^TN\tilde{M} \biggr) \\
%& = -\frac{1}{2} \biggl(\gammastar^T H \gammastar-\gammastar^T HJ -J^T H \gammastar + (c_t-D-E)^TM(c_t-D-E)  +\tilde{M}^TN\tilde{M}\biggr)\\
&= -\frac{1}{2} \biggl( (\gammastar-J_t)^\top H_t (\gammastar-J_t)-J_t^\top H_t J_t + (c_t-D-E_t)^\top M(c_t-D-E_t)  +\tilde{M}_t^\top N_t \tilde{M}_t \biggr),
\end{align} where  $A_t$ is defined in \eqref{eq:A}, $M$ is defined in \eqref{eq:M}, $\tilde{M}_t$ in \eqref{eq:tildeM}, $N$ in \eqref{eq:N},  $D= (M^{-1})^\top \Sigma_c^{-1}\mu_c$, $E_t=(M^{-1})^\top \Sigma_n^{-1}\chat_t$,
$H_t = \Sigma_n^{-1}(M^{-1})^\top \Sigma_n^{-1}+N_t$ and $J_t=(H^{-1}_t)^\top \Bigl(\Sigma_n^{-1}(-c_t+D+E_t)+N_t\tilde{M}_t\Bigr)$.

Subsequently, we have
\begin{align}
    \log p(c_t|\chat_t,\Hscr_{t-1,\chat}) &\propto  -\frac{1}{2} \biggl(-J_t^\top H_t J_t + (c_t-D-E_t)^\top M(c_t-D-E_t) \biggr)\\
 %   &=-\frac{1}{2} \biggl(-(-c_t^T\Sigma_n^{-1}+L)(H^{-1})^T(-\Sigma_n^{-1}c_t+L^T) + (c_t-D-E)^TM(c_t-D-E) \biggr)\\
    & \propto -\frac{1}{2} \biggl( c_t^\top \Bigl(M-\Sigma_n^{-1}(H_t^{-1})^\top \Sigma_n^{-1} \Bigr)c_t -c_t^\top \Bigl(M(D+E_t)-\Sigma_n^{-1}(H_t^{-1})^\top L_t^\top \Bigr) \nonumber \\& - \Bigl((D+E_t)^\top M -L_t(H_t^{-1})\Sigma_n^{-1}\Bigr),
\end{align}where $L_t =(D+E_t)^\top \Sigma_n^{-1}+\tilde{M}_t^\top N_t$.
Thus, we have,
\begin{align}
     P(c_t|\chat_t,\Hscr_{t-1,\chat}) &= \Nscr(c_t|V_t, R_t^{-1}), \quad \mbox{where} \\
    & R_t = M-\Sigma_n^{-1}(H_t^{-1})^\top \Sigma_n^{-1} \\
    & V_t = (R_t^{-1})^\top \Bigl(M(D+E_t)-\Sigma_n^{-1}(H_t^{-1})^\top L_t^\top \Bigr) .
    \end{align}
 
\subsection{Analytical Tractability of the True Posterior Distribution $P(\thetastar|\Hscr_{t-1,r,a,\chat})$}\label{app:analyticaltractability}
In this subsection, we discuss the difficulty in evaluating the exact posterior distribution $P(\thetastar|\Hscr_{t-1,r,a,\chat})$. To see this, note that the true posterior distribution can be obtained via the Baye's theorem as
\begin{align}P(\thetastar|\Hscr_{t-1,r,a,\chat}) \propto P(\thetastar) \prod_{\tau=1}^{t-1} P(r_{\tau}|a_{\tau},\chat_{\tau},\Hscr_{\tau-1,\chat}), \label{eq:1111}\end{align}
where \begin{align}P(r_{\tau}|a_{\tau},\chat_{\tau},\Hscr_{\tau-1,\chat}) =\Ebb_{P(c_{\tau}|\chat_{\tau},\Hscr_{\tau-1,\chat})}[P(r_{\tau}|a_{\tau},c_{\tau})]\end{align} is obtained by averaging the reward distribution $P(r_{\tau}|a_{\tau},c_{\tau})$ with respect to the predictive distribution $P(c_{\tau}|\chat_{\tau},\Hscr_{\tau-1,\chat})$. When the mean reward function is defined using general feature maps, the resulting distribution need not be Gaussian. 

Now assume the following feature map $\phi(a,c)=G(a)c$ where $G(a)$ is a $d \times d$ transformation matrix satisfying Assumption~\ref{assum:1}. Using the above linear feature map (linear in $c$), we get that $P(r_t|a_t,\chat_t,\Hscr_{t-1,\chat}) =\Nscr(\tilde{m}_t,\nu_t)$ is Gaussian with mean \begin{align}\tilde{m}_t=\thetastart G(a)\Ebb_{P(c_t|\chat_t,\Hscr_{t-1,\chat})}[c_t] \label{eq:mean_1} \end{align} and variance \begin{align}\nu_t=\thetastart {\rm var}(c_t)\thetastar+\sigma^2, \label{eq:variance_1}\end{align} with ${\rm var}(c_t)=G(a)\Ebb_{P(c_t|\chat_t,\Hscr_{t-1,\chat})}[c_tc_t^\top]G(a)^\top -G(a)(\Ebb_{P(c_t|\chat_t,\Hscr_{t-1,\chat})}[c_t])(\Ebb_{P(c_t|\chat_t,\Hscr_{t-1,\chat})}[c_t])^\top G(a)^\top$. Note the dependence of the variance $\nu_t$ on $\thetastar$. Consequently, it can be verified that  plugging the distribution $P(r_t|a_t,\chat_t,\Hscr_{t-1,\chat})$ in Bayes theorem of  \eqref{eq:1111} does not give a tractable Gaussian posterior distribution.
\subsection{Proof of Lemma~\ref{lem:r_CB_withcontextdistributions}}
We start by distinguishing the true and approximated posterior distributions. Recall that $P_t(\thetastar):=P(\thetastar|\Hscr_{t-1,r,a,\chat})$ denotes the true posterior and $\bar{P}_t(\thetastar):=\bar{P}(\thetastar|\Hscr_{t-1,r,a,\chat})$ denotes the approximated posterior. We then denote $P_t(\hat{a}_t,\thetastar):=P(\hat{a}_t,\thetastar|\mathcal{F}_t)$ as the  distribution of $\hat{a}_t$ and $\thetastar$ conditioned on $\Fcal_t$, while $\bar{P}_t(\hat{a}_t,\thetastar):=\bar{P}(\hat{a}_t,\thetastar|\mathcal{F}_t)$ denote the distribution of $\hat{a}_t$ and $\thetastar$ under the sampling distribution. Furthermore, we have that $\bar{P}_t(a_t,\thetastar)=\bar{P}_t(a_t)\bar{P}_t(\thetastar)=P_t(a_t)\bar{P}_t(\thetastar)$. We start by decomposing $\Rcal^T_{\rm CB}$ into the following three differences,
\begin{align}
\Rcal^T_{\rm CB}&=\sum_{t=1}^T \Ebb \biggl[\Ebb_{P_t(\hat{a}_t,\thetastar)}[\psi(\hat{a}_t)^\top\thetastar] - \Ebb_{P_t(a_t,\thetastar)}[\psi(a_t)^\top\thetastar]\biggr] \nonumber \\
&=\sum_{t=1}^T \Ebb \biggl[ \underbrace{\Ebb_{\bar{P}_t(\hat{a}_t,\thetastar)}[\psi(\hat{a}_t)^\top\thetastar] - \Ebb_{\bar{P}_t(a_t,\thetastar)}[\psi(a_t)^\top\thetastar]}_{:=\Tone} +\underbrace{\Ebb_{P_t(\hat{a}_t,\thetastar)}[\psi(\hat{a}_t)^\top\thetastar] - \Ebb_{\bar{P}_t(\hat{a}_t,\thetastar)}[\psi(\hat{a}_t)^\top\thetastar]] }_{:=\Ttwo} \nonumber \\& + \underbrace{\Ebb_{\bar{P}_t(a_t,\thetastar)}[\psi(a_t)^\top\thetastar]-\Ebb_{P_t(a_t,\thetastar)}[\psi(a_t)^\top\thetastar]\biggr]}_{:=\Tthree} 
\end{align}
$\Tone$ can be upper bounded similar to Lemma~\ref{lem:R_CB_delayed}, and we will detail this. 

\subsubsection*{Upper Bound on $\Ttwo$} To obtain an upper bound on $\Ttwo$, note that the following equivalence holds  $\Ebb_{P_t(\hat{a}_t,\thetastar)}[\psi(\hat{a}_t)^\top\thetastar]=\Ebb_{{P}_t(\thetastar)}\Bigl[\max_{a} \psi(a)^\top\thetastar \Bigr]$. Using this, we can rewrite $\Ttwo$ as
\begin{align}
    \Ttwo= \Ebb_{P_t(\thetastar)}[\max_a \psi(a)^\top\thetastar] - \Ebb_{\bar{P}_t(\thetastar)}[\max_a \psi(a)^\top\thetastar].
\end{align} Note here that when $\thetastar \sim \bar{P}_t(\thetastar)$, for each $a \in \Acal$, we have that $z_a=\psi(a)^\top\thetastar$ follows Gaussian distribution $\Nscr(z_a|\psi(a)^\top \mu_{t-1}, \psi(a)^\top\Sigma_{t-1}^{-1}\psi(a))$ with mean $\psi(a)^\top \mu_{t-1}$ and variance $\psi(a)^\top\Sigma_{t-1}^{-1}\psi(a)$, where $\mu_{t-1}$ and $\Sigma_{t-1}$ are as defined in \eqref{eq:mean_theta} and \eqref{eq:variance_theta} respectively. Thus, $\Ebb_{\bar{P}_t(\thetastar)}[\max_a z_a]$ is the average of maximum of Gaussian random variables. We can then apply Lemma~\ref{lem:maxgaussian_com} with $P(x) =P_t(\thetastar)$, $Q(x)=\bar{P}_t(\thetastar)$, $n=|\Acal|=K$, $\mu_i = \psi(a)^\top\mu_{t-1}$ and $\sigma_i=\psi(a)^\top \Sigma_{t-1}^{-1}\psi(a)$ to get that
\begin{align}
 \Ttwo&\leq    \sqrt{2 (\log K+D_{\rm KL}(P_t(\thetastar) \Vert \bar{P}_t(\thetastar))\max_a \psi(a)^\top \Sigma_t^{-1}\psi(a)}.
\end{align} Using this, we get that
\begin{align}
    \sum_{t=1}^T\Ebb[\Ttwo] &\leq \Ebb\biggl[ \sum_{t=1}^T \sqrt{2 (\log K+D_{\rm KL}(P_t(\thetastar) \Vert \bar{P}_t(\thetastar)))\max_a \psi(a)^\top \Sigma_t^{-1}\psi(a)} \biggr] \nonumber \\
    & \leq \sqrt{\Bigl( \sum_{t=1}^T \Ebb\Bigl[2 (\log K+D_{\rm KL}(P_t(\thetastar) \Vert \bar{P}_t(\thetastar)))\Bigr]\Bigr)\Bigl( \sum_{t=1}^T \Ebb\Bigl[\max_a \psi(a)^\top \Sigma_t^{-1}\psi(a)\Bigr]\Bigr)}, \label{eq:T2_eq2}
\end{align}where the last inequality follows from an application of Cauchy-Schwarz inequality. We now upper bound each of the two terms in \eqref{eq:T2_eq2}. 

We first note that 
\begin{align}
    \sum_{t=1}^T \Ebb \Bigl[\max_a \psi(a)^\top \Sigma_t^{-1} \psi(a)\Bigr] \leq  \sum_{t=1}^T \Ebb \Bigl[\max_a \psi(a)^\top (\lambda \mathbb{I}) \psi(a)\Bigr] \leq \lambda T. \label{eq:22}
\end{align}where the first inequality follows since $\Sigma_t^{-1} \leq \lambda \Ibb$.

 When feature map $\phi(a,c)=G_a c$, it is easy to verify that $P(r_t|a_t,\Hscr_{t,\chat},\thetastar) =\Nscr(\tilde{m}_t,\nu_t)$ where the mean and variance are given as in \eqref{eq:mean_1} and \eqref{eq:variance_1}. Furthermore, we consider the distribution $\bar{P}(r_t|a_t,\Hscr_{t,\chat},\thetastar)=\Nscr(r_t|\tilde{m}_t,\sigma^2)$. Subsequently, we define
 $$P_t(\thetastar)P(\Fcal_t)=P(\thetastar)P(\mathcal{F}_t|\thetastar)=P(\thetastar)P(\Hscr_{t-1,r}|\Hscr_{t-1,a,\chat},\thetastar) P(\Hscr_{t-1,a,\chat},\chat_t)$$ with  $$P(\Hscr_{t-1,r}|\Hscr_{t-1,a,\chat},\thetastar) =\prod_{\tau=1}^{t-1}P(r_{\tau}|a_{\tau},\Hscr_{{\tau},\chat},\thetastar).$$  Similarly, it is easy to verify that $\bar{P}_t(\thetastar) \propto P(\thetastar) \prod_{\tau=1}^{t-1} \bar{P}(r_{\tau}|a_{\tau},\Hscr_{\tau,\chat},\thetastar)$ whereby we get that $$\bar{P}_t(\thetastar)\bar{P}(\Fcal_t)=P(\thetastar)\bar{P}(\mathcal{F}_t|\thetastar)=P(\thetastar)\bar{P}(\Hscr_{t-1,r}|\Hscr_{t-1,a,\chat},\thetastar) P(\Hscr_{t-1,a,\chat},\chat_t)$$ with  $\bar{P}(\Hscr_{t-1,r}|\Hscr_{t-1,a,\chat},\thetastar) =\prod_{\tau=1}^{t-1}\bar{P}(r_{\tau}|a_{\tau},\Hscr_{{\tau},\chat},\thetastar) $. 
 
 Using all these, we can upper bound the KL divergence term as follows:
\begin{align}
    D_{\rm KL}(P_t(\thetastar) \Vert \bar{P}_t(\thetastar)) &\leq D_{\rm KL}(P_t(\thetastar)P(\Fcal_t) \Vert \bar{P}_t(\thetastar)\bar{P}(\Fcal_t)) \nonumber \\
    & = \Ebb_{P(\thetastar)P(\Hscr_{t-1,a,\chat},\chat_t)}\Bigl[D_{\rm KL} (P(\Hscr_{t-1,r}|\Hscr_{t-1,a,\chat},\chat_t,\thetastar) \Vert \bar{P}(\Hscr_{t-1,r}|\Hscr_{t-1,a,\chat},\chat_t,\thetastar) \Bigr]\nonumber \\
    & =\sum_{\tau=1}^{t-1} \Ebb_{P(\thetastar)P(\Hscr_{t-1,a,\chat},\chat_t)}\Bigl[D_{\rm KL}(P(r_{\tau}|a_{\tau},\Hscr_{{\tau},\chat},\thetastar) \Vert \bar{P}(r_{\tau}|a_{\tau},\Hscr_{{\tau},\chat},\thetastar))\Bigr]\nonumber \\
    & = \sum_{\tau=1}^{t-1} \Ebb_{P(\thetastar)P(\Hscr_{t-1,a,\chat},\chat_t)}\Bigl[\frac{1}{2}\log \frac{\sigma^2}{\sigma^2+\thetastart {\rm var}(c_{\tau})\thetastar} +\frac{1}{2} \frac{\sigma^2+\thetastart {\rm var}(c_{\tau})\thetastar}{\sigma^2 }-\frac{1}{2} \Bigr]\nonumber \\
    & \leq \sum_{\tau=1}^{t-1} \Ebb_{P(\thetastar)P(\Hscr_{t-1,a,\chat},\chat_t)}\Bigl[\frac{1}{2} \frac{\thetastart {\rm var}(c_{\tau})\thetastar}{\sigma^2}\Bigr] \\
    & =\sum_{\tau=1}^{t-1} \Ebb_{P(\Hscr_{t-1,a,\chat},\chat_t)}\Bigl[\frac{{\rm Tr}\Bigl({\rm var}(c_{\tau})\Ebb_{P(\thetastar)}[\thetastar \thetastart]\Bigr) }{2\sigma^2}\Bigr]\nonumber \\
    & = \frac{\lambda \sum_{\tau=1}^{t-1} \Ebb[{\rm Tr}({\rm var}(c_{\tau}))] }{2\sigma^2} \leq \frac{\lambda}{2\sigma^2}(t-1),
\end{align} where the last inequality follows by noting that ${\rm Tr}({\rm var}(c_{\tau}))\leq 1$.
Thus, we get that 
\begin{align}
    \sum_{t=1}^T \Ebb[D_{\rm KL}(P_t(\thetastar) \Vert \bar{P}_t(\thetastar)) ] \leq  \frac{\lambda}{4\sigma^2}(T^2-T)\leq \frac{\lambda T^2}{4\sigma^2}\label{eq:11}.
\end{align}

Using \eqref{eq:11} and \eqref{eq:22} in \eqref{eq:T2_eq2}, we get that 
\begin{align}
    \sum_{t=1}^T \Ebb[\Ttwo] &\leq   \sqrt{2\lambda T\Bigl( \sum_{t=1}^T \Ebb\Bigl[ (\log K+D_{\rm KL}(P_t(\thetastar) \Vert \bar{P}_t(\thetastar)))\Bigr]\Bigr)},  \\ &= \sqrt{2\lambda T\Bigl( T\log K+\frac{\lambda T^2}{4\sigma^2}\Bigr)}.
\end{align}
 If $\lambda \leq \frac{d \sigma^2}{T}$, we get that 
\begin{align}
     \sum_{t=1}^T \Ebb[\Ttwo] \leq \sqrt{2d \sigma^2(T \log K+ dT/4)}.
\end{align}
\subsubsection*{Upper Bound on $\Tthree$}
We can bound $\Tthree$ by observing that 
\begin{align}
\Tthree & = \Ebb_{P_t(a_t)}[\psi(a_t)]^\top \Bigl(\Ebb_{\bar{P}_t(\thetastar)}[\thetastar] -\Ebb_{P_t(\thetastar)}[\thetastar] \Bigr) \\
& = \Ebb_{\bar{P}_t(\thetastar)}[\Psi_t^\top \thetastar] - \Ebb_{P_t(\thetastar)}[\Psi_t^\top \thetastar]
\end{align} where we used $\Psi_t =\Ebb_{P_t(a_t)}[\psi(a_t)] $. Note that for $\thetastar \sim \bar{P}_t(\thetastar)$, the random variable $\Psi_t^\top \thetastar$ is Gaussian with mean $\Psi_t^\top \mu_{t-1}$ and variance $\Psi_t^\top \Sigma_{t-1}^{-1}\Psi_t$.
Consequently, $\Psi_t^\top \thetastar$ is also $\Psi_t^\top \Sigma_{t-1}^{-1}\Psi_t$-sub-Gaussian according to Definition A.1. By using Lemma~\ref{lem:DVinequality}, we then get that
\begin{align}
    |\Tthree| \leq \sqrt{2 (\Psi_t^\top \Sigma_t^{-1}\Psi_t) D_{\rm KL}(P_t(\thetastar) \Vert \bar{P}_t(\thetastar)) }.
\end{align}
Using Cauchy-Schwarz inequality then yields that
\begin{align}
    \Ebb[\sum_{t=1}^T |\Tthree|] & \leq \sqrt{\Bigl(\sum_{t=1}^T \Ebb[\Psi_t^\top \Sigma_t^{-1}\Psi_t] \Bigr) \Bigl(\sum_{t=1}^T \Ebb[2D_{\rm KL}(P_t(\thetastar) \Vert \bar{P}_t(\thetastar)) ] \Bigr)} \nonumber \\
    & \leq \sqrt{2\lambda T \frac{\lambda T^2}{4 \sigma^2}} \leq \sqrt{d^2 T\sigma^4/2}.
\end{align}where the second inequality follows from \eqref{eq:11} and \eqref{eq:22}.

\subsubsection*{Upper Bound on $\Tone$}
Note that in $\Tone$, $\bar{P}_t(\hat{a}_t) =\bar{P}_t(a_t) =P_t(a_t)$, whereby the posterior is matched. Hence, one can apply bounds from conventional contextual Thompson Sampling here.
For simplicity, we denote $\bar{\Ebb}_t[\cdot] = \Ebb_{\bar{P}}[\cdot|\mathcal{F}_t]$ to denote the expectation with respect to $\bar{P}_t(a,\theta)$. To this end, as in the proof of Lemma~\ref{lem:R_CB_delayed}, we start by defining an information ratio,
\begin{align}
    \Gamma_t = \frac{\Tone^2}{\bar{E}_t\Bigl[\Bigl(\psi(a_t)^\top\thetastar - \psi(a_t)^\top\mu_t)\Bigr)^2\Bigr]:=\Lambda_t}, 
\end{align}using which we get the upper bound on $\Tone$ as
\begin{align}
    \sum_{t=1}^T\Ebb[\Tone] \leq \Ebb \Bigl[\sum_{t=1}^T \sqrt{\Gamma_t\Lambda_t}\Bigr] \leq \sqrt { \Bigl(\sum_{t=1}^T \Ebb[\Gamma_t] \Bigr)\Bigl(\sum_{t=1}^T \Ebb[\Lambda_t] \Bigr)}
\end{align}by the Cauchy-Schwarz inequality. 

Furthermore, we have that
\begin{align}
    \Lambda_t = \bar{\Ebb}_t \Bigl[ (\psi(a_t)^\top(\thetastar-\mu_t))^2\Bigr] &= \bar{\Ebb}_t \Bigl[\psi(a_t)^\top (\thetastar-\mu_{t-1})(\thetastar-\mu_{t-1})^\top \psi(a_t)\Bigr] \nonumber \\
&=\bar{\Ebb}_t[\psi(a_t)^\top\Sigma_{t-1}^{-1}\psi(a_t)] =\bar{E}_t[\Vert \psi(a_t)\Vert_{\Sigma_{t-1}^{-1}}],
\end{align}where $\mu_{t-1}$ and $\Sigma_{t-1}$ are defined as in \eqref{eq:mean_theta} andd \eqref{eq:variance_theta}. Subsequently, using elliptical potential lemma, we get that
\begin{align}
\sum_{t=1}^T \Lambda_t \leq 2m \sigma^2 \log \Bigl( 1+\frac{(T)\lambda}{m \sigma^2}\Bigr). \label{eq:T1_1}
\end{align}

To obtain an upper bound on the information ratio $\Gamma_t$, we define $\bar{f}(\thetastar,a)=\psi(a)^\top\thetastar$ and $\bar{f}(a)=\psi(a)^\top\mu_{t-1}$ and let
\begin{align}
    M_{a,a'} = \sum_{a,a'}\sqrt{\bar{P}_t(a_t=a)\bar{P}_t(\hat{a}_t=a')} (\bar{\Ebb}_t[\bar{f}(\thetastar,a)|\hat{a}_t=a']-\bar{f}(a)).
\end{align}
It is easy to see that
\begin{align}
    \Tone &= \bar{\Ebb}_t[\bar{f}(\hat{a}_t,\thetastar)]-\bar{\Ebb}_t[\bar{f}(a_t)] \nonumber \\
    & = \sum_{a'} \bar{P}_t(\hat{a}_t=a')\Bigl(\bar{\Ebb}_t[\bar{f}(\thetastar,a')|\hat{a}_t=a'] - \bar{\Ebb}_t[\bar{f}(\hat{a}_t)]\Bigr)\nonumber \\
    & = \sum_{a'} \bar{P}_t(\hat{a}_t=a') \Bigl(\bar{\Ebb}_t[\bar{f}(\thetastar,a')|\hat{a}_t=a'] - \bar{f}(a') \Bigr)={\rm Tr}(M),
\end{align}where the second  and last equality follows since $\bar{P}_t(a_t)=\bar{P}_t(\hat{a}_t)$. Similarly, we can relate $\Lambda_t$ with the matrix $(M_{a,a'})$ as
\begin{align}
\Lambda_t & = \bar{\Ebb}_t[(\bar{f}(\thetastar,a_t)-\bar{f}(a_t))^2] \\
& =\sum_a \bar{P}_t(a_t=a) \bar{\Ebb}_t[(\bar{f}(\thetastar,a)-\bar{f}(a))^2]\nonumber \\
& = \sum_{a,a'}\bar{P}_t(a_t=a) \bar{P}_t(\hat{a}_t=a')\bar{\Ebb}_t[(\bar{f}(\thetastar,a)-\bar{f}(a))^2|\hat{a}_t=a'] \nonumber \\
& \geq \sum_{a,a'}\bar{P}_t(a_t=a) \bar{P}_t(\hat{a}_t=a')\Bigl(\bar{\Ebb}_t[(\bar{f}(\thetastar,a)-\bar{f}(a))|\hat{a}_t=a']\Bigr)^2 \nonumber \\
& =\sum_{a,a'}\bar{P}_t(a_t=a) \bar{P}_t(\hat{a}_t=a')\Bigl(\bar{\Ebb}_t[(\bar{f}(\thetastar,a)|\hat{a}_t=a']-\bar{f}(a)\Bigr)^2 =\Vert M \Vert_F^2 ,
\end{align} whereby we get that
\begin{align}
    \Gamma_t \leq \frac{{\rm Tr}(M)^2}{\Vert M \Vert_F^2}  \leq m,
\end{align} where the last inequality can be proved as in \citep{russo2016information}. Following \citep{neu2022lifting}, it can be seen that $\Lambda_t \leq 2 \log(1+K)$ also holds. Using this, together with the upper bound \eqref{eq:T1_1} gives that 
\begin{align}
    \sum_{t=1}^T \Ebb[\Tone] \leq \sqrt{2Tm\sigma^2 \min\{m,2\log(1+K)\} \log \Bigl( 1+\frac{T\lambda}{m\sigma^2}\Bigr)}.
\end{align}

\subsection{Proof of Lemma~\ref{lem:estmationerror_nocontext}}
We first give an upper bound on the estimation error that does not require the assumption of a linear feature map.
\paragraph{A General Upper Bound on $\Rcal^T_{\rm EE1}$:}\label{app:generalupperbound}
To get an upper bound on $\Rcal^T_{\rm EE1}$ that does not require the assumption that $\phi(a,c)=G(a)c$, we leverage 
the same analysis as in the proof of Lemma~\ref{lem:EE_1}. Subsequently, we get that
\begin{align}\Rcal^T_{\rm{EE1}}&\leq \sum_{t=1}^T \Ebb \Bigl[ \Delta\Bigl(P(c_t|\chat_t,\gammastar), P(c_t|\chat_t,\Hscr_{\chat})\Bigr) \indicator\{\Escr\}\Bigr] + 2\delta T \Ebb[\Vert \thetastar \Vert_2 |\Escr^c] \non \\
& \leq \sum_{t=1}^T \Ebb \Bigl[ \Delta\Bigl(P(c_t|\chat_t,\gammastar), P(c_t|\chat_t,\Hscr_{\chat})\Bigr) \indicator\{\Escr\}\Bigr] + 2\delta^2 T \sqrt{\frac{m\lambda}{2\pi}}, \non
\end{align}where the event $\Escr$ is defined as in \eqref{eq:event}. Subsequently, the first summation can be upper bounded as
\begin{align}
   \sum_{t=1}^T  \Ebb \Bigl[ \Delta\Bigl(P(c_t|\chat_t,\gammastar), P(c_t|\chat_t,\Hscr_{\chat})\Bigr) \indicator\{\Escr\}] &\leq \sqrt{2TU^2 \sum_{t=1}^T I(c_t;\gammastar|\chat_t,\Hcal_{\chat})} 
   \label{eq:EE1_ub_app_nocontext} \non\\
   & = \sqrt{2TU^2 \sum_{t=1}^T (H(c_t|\chat_t,\Hscr_{\chat})-H(c_t|\chat_t,\gammastar))} \non \\
   & = \sqrt{TU^2 \sum_{t=1}^T \log \Bigl( {\rm det}(R_t^{-1}) {\rm det}(M_t)\Bigr)} \non\\
   & \leq \sqrt{TU^2 \frac{d\sigma^2_c}{\sigma^2_n} \Bigl(\frac{\sigma^2_{\gamma}}{\sigma^2_c+\sigma^2_n}+ \log (T-1)\Bigr) } \non
\end{align} where $U$ is defined as in \eqref{eq:event},  $M_t =\Sigma_c^{-1}+\Sigma_n^{-1}$ and $R_t$ is as in \eqref{eq:variance_predcontext_denoising}. The last inequality is derived in Section~\ref{sec:R_tanalysis} using that $\Sigma_c =\sigma^2_c \Ibb$, $\Sigma_n = \sigma^2_n \Ibb$ and $\Sigma_{\gamma}=\sigma^2_{\gamma}\Ibb$.

We thus get that
\begin{align}
    \Rcal^T_{\rm EE1} \leq \sqrt{TU^2 \frac{d\sigma^2_c}{\sigma^2_n} \Bigl(\frac{\sigma^2_{\gamma}}{\sigma^2_c+\sigma^2_n}+ \log (T-1)\Bigr)} +2\delta^2T \sqrt{\frac{m\lambda}{2\pi}}
\end{align}for $\delta \in (0,1)$.
We note that same upper bound holds for the term $\Rcal^T_{\rm EE2}$.

\paragraph{Upper Bound for linear feature maps:} We now obtain an upper bound on the estimation error under the assumption of a linear feature map $\phi(a,c)=G_a c$ such that $\Vert \phi(a,c) \Vert_2 \leq 1$. The following set of inequalities hold:
\begin{align}\Rcal^T_{\rm{EE1}}&=\sum_{t=1}^T \Ebb \Bigl[\psi(a^{*}_t,\chat_t|\gammastar)^\top \thetastar -\psi(\hat{a}_t,\chat_t|\Hscr_{\chat})^\top\thetastar \Bigr] \nonumber \\& \leq  \sum_{t=1}^T \Ebb \Bigl[\psi(a^{*}_t,\chat_t|\gammastar)^\top \thetastar -\psi(a^{*}_t,\chat_t|\Hscr_{\chat})^\top\thetastar \Bigr] \nonumber \\
& = \sum_{t=1}^T \Ebb \Bigl[\Ebb_{P(c_t|\chat_t,\gammastar)}[\phi(a^{*}_t,c_t)^\top \thetastar] -\Ebb_{P(c_t|\chat_t,\Hscr_{\chat})}[\phi(a^{*}_t,c_t)^\top\thetastar ] \label{eq:T1_2}.
\end{align}
Note that $P(c_t|\chat_t,\Hscr_{\chat}) = \Nscr(c_t|V_t,R_t^{-1})$ where $R_t$ and $V_t$ are respectively defined in \eqref{eq:variance_predcontext_denoising} and \eqref{eq:mean_predcontext_denoising}.
Consequently, $\phi(a^*_t,c_t)^{\top} \thetastar=c_t^{\top} G_{a^*_t}^{\top} \thetastar$
is $ s_t^2=\theta^{*\top} G_{a^*_t} R_t^{-1}G_{a^*_t}^{\top} \thetastar$-sub-Gaussian with respect to 
$P(c_t|\chat_t,\Hscr_{\chat})$. Consequently, using Lemma~\ref{lem:DVinequality}, we can upper bound the inner expectation of \eqref{eq:T1_2} as
\begin{align}
    |\Ebb_{P(c_t|\chat_t,\gammastar)}[\phi(a^{*}_t,c_t)^\top \thetastar] -\Ebb_{P(c_t|\chat_t,\Hscr_{\chat})}[\phi(a^{*}_t,c_t)^\top\thetastar| \leq \sqrt{2s_t^2 D_{\rm KL}(P(c_t|\chat_t,\gammastar)\Vert P(c_t|\chat_t,\Hscr_{\chat}))}.
\end{align}Summing over $t$ and using Cauchy-Schwarz inequality then gives that
\begin{align}
    \Rcal^T_{\rm{EE1}}& \leq \sqrt{2 (\sum_{t=1}^T \Ebb[s^2_t]) \Bigl(\sum_{t=1}^T \Ebb\Bigl[D_{\rm KL}(P(c_t|\chat_t,\gammastar)\Vert P(c_t|\chat_t,\Hscr_{\chat})) \Bigr] \Bigr)}. \label{eq:REE1}
\end{align}
We now evaluate the KL-divergence term. To this end, note that conditioned on $\gammastar$ and $\chat_t$, $c_t$ is independent of $\Hscr_{\chat}$, \emph{i.e.,}, $P(c_t|\chat_t,\gammastar,\Hscr_{\chat})=P(c_t|\chat_t,\gammastar)$. This gives that
\begin{align}
    \sum_{t=1}^T \Ebb\Bigl[D_{\rm KL}(P(c_t|\chat_t,\gammastar)\Vert P(c_t|\chat_t,\Hscr_{\chat})) \Bigr] &= \sum_{t=1}^T I(c_t;\gammastar|\chat_t,\Hscr_{\chat}) \nonumber \\
    & = \sum_{t=1}^T H(c_t|\chat_t,\Hscr_{\chat}) - H(c_t|\chat_t,\gammastar) \nonumber \\
    &=\frac{1}{2}\sum_{t=1}^T  \Ebb_{P(\chat_t,\Hscr_{\chat})}[\log {\rm det}(R^{-1}_t)]-\frac{1}{2} \sum_{t=1}^T \Ebb_{P(\chat_t,\gammastar)}[\log {\rm det}(M^{-1}_t)] \nonumber \\
    & = \sum_{t=1}^T \frac{1}{2} \Ebb_{P(\chat_t,\Hscr_{\chat})}[\log \Bigl({\rm det}(R^{-1}_t) {\rm det}(M_t)\Bigr) \nonumber \\
    & \leq \frac{d\sigma^2_c}{\sigma^2_n} \Bigl(\frac{\sigma^2_{\gamma}}{\sigma^2_n+\sigma^2_c}+ \log (T-1)\Bigr),\label{eq:REE1_1}
\end{align}where $M_t =\Sigma_c^{-1}+\Sigma_n^{-1}$ and $R_t$ is as in \eqref{eq:variance_predcontext_denoising}. The first equality follows by noting that $\Ebb\Bigl[D_{\rm KL}(P(c_t|\chat_t,\gammastar)\Vert P(c_t|\chat_t,\Hscr_{\chat})) \Bigr]=\Ebb_{P(\chat_t,\Hscr_{\chat})}\Bigl[\Ebb_{P(\gammastar|\chat_t,\Hscr_{\chat})}\Bigl[D_{\rm KL}(P(c_t|\chat_t,\gammastar,\Hscr_{\chat})\Vert P(c_t|\chat_t,\Hscr_{\chat})) \Bigr]\Bigr]=\Ebb_{P(\chat_t,\Hscr_{\chat})}[I(c_t;\gammastar|\chat_t,\Hscr_{\chat})]$ with the outer expectation taken over $\chat_t$ and $\Hscr_{\chat}$. The last inequality is proved in Section~\ref{sec:R_tanalysis} using that $\Sigma_c =\sigma^2_c \Ibb$, $\Sigma_n = \sigma^2_n \Ibb$ and $\Sigma_{\gamma}=\sigma^2_{\gamma}\Ibb$.

We can now upper bound $\sum_{t}\Ebb[s^2_t]$ as follows.
\begin{align*}
  \Ebb[s^2_t] &\leq \Ebb[\max_a \theta^{*\top} G_a R_t^{-1}G_a^\top \thetastar]  \leq \sum_a \Ebb[\theta^{*\top} G_a R_t^{-1}G_a^\top\thetastar] = \sum_a {\rm Tr}(G_a R_t^{-1}G_a^\top\Ebb[\thetastar \theta^{*\top}]) =\lambda \sum_a {\rm Tr}(R_t^{-1}G_a^\top G_a) \nonumber \\
  & = \lambda b_t {\rm Tr}(\sum_a G_a^\top G_a)
\end{align*} where the last equality uses that $R_t^{-1}=b_t\Ibb$ as in \eqref{eq:Rtinv}. 
Using \eqref{eq:Rtinv}, we get that 
\begin{align*}
    \sum_t b_t &= \frac{\sigma^2_c \sigma^2_n}{f} \sum_t\Bigl(1+\frac{\sigma^2_c \sigma^2_{\gamma}}{(t-1)\sigma^2_{\gamma}\sigma^2_n +f \sigma^2_n}\Bigr)\nonumber \\
    & =\frac{\sigma^2_c \sigma^2_n T}{f} + \sum_t \frac{\sigma^2_c }{f} \frac{\sigma^2_c \sigma^2_{\gamma}}{(t-1)\sigma^2_{\gamma} +f }\\
    & =\frac{\sigma^2_c \sigma^2_n T}{f} + \frac{\sigma^4_c \sigma^2_{\gamma}}{f^2 }+\sum_{t>1} \frac{\sigma^2_c }{f} \frac{\sigma^2_c \sigma^2_{\gamma}}{(t-1)\sigma^2_{\gamma} +f }\\
    &  \leq \frac{\sigma^2_c \sigma^2_n}{f}T + \frac{\sigma^4_c \sigma^2_{\gamma}}{f^2 }+\frac{\sigma^4_c }{f} \log(T-1).
\end{align*}Using the above relation, we get that
\begin{align*}
    \sum_t \Ebb[s^2_t]=\lambda  {\rm Tr}(\sum_a G_a^\top G_a) \sum_t b_t \leq \frac{\lambda K\sigma^2_c}{f}    \max_a{\rm Tr}( G_a^\top G_a) \Bigl(  \sigma^2_n T + \frac{\sigma^2_c \sigma^2_{\gamma}}{f}+ \sigma^2_c \log(T-1)\Bigr).
\end{align*}
 If $\lambda \leq \frac{d \sigma^2}{T}$, we get that
\begin{align*}
  \sum_t \Ebb[s^2_t] \leq  \frac{d\sigma^2 K\sigma^2_c}{f}  \max_a  {\rm Tr}( G_a^\top G_a) \Bigl(  \sigma^2_n + \frac{\sigma^2_c \sigma^2_{\gamma}}{Tf}+\sigma^2_c \frac{\log(T-1)}{T}\Bigr)=L
\end{align*} Using the above inequality together with \eqref{eq:REE1_1} in \eqref{eq:REE1} yields that
\begin{align}
    \Rcal^T_{{\rm EE1}} \leq \sqrt{2L\frac{d\sigma^2_c}{\sigma^2_n} \Bigl(\frac{\sigma^2_{\gamma}}{\sigma^2_n+\sigma^2_c}+ \log (T-1)\Bigr)}.
\end{align}
An upper bound on $\Rcal^T_{{\rm EE2}}$ similarly follows.
\subsubsection{Analysis of $R_t$}\label{sec:R_tanalysis} Assume that $\Sigma_n=\sigma^2_n \Ibb$, $\Sigma_c =\sigma^2_c \Ibb$ and $\Sigma_{\gamma}=\sigma^2_{\gamma} \Ibb$. Then, from \eqref{eq:variance_predcontext_denoising}, we get that
\begin{align}
H_t &= (t-1)\Sigma_n^{-1}-(t-2)\Sigma_n^{-1}M^{-1}\Sigma_n^{-1}+\Sigma_{\gamma}^{-1} \nonumber \\
& = \Bigl(\frac{(t-1)}{\sigma_n^2} -\frac{(t-2)\sigma_n^2 \sigma_c^2}{\sigma^4_n\underbrace{(\sigma^2_c+\sigma^2_n)}_{:=f}} + \frac{1}{\sigma_{\gamma}^2}\Bigr) \Ibb \nonumber  \\
%& =\Bigl(\frac{(t-1)}{\sigma_n^2} -\frac{(t-2) \sigma_c^2}{a \sigma^2_n} + \frac{1}{\sigma_{\gamma}^2}\Bigr) \Ibb \\
& = \frac{(t-1)\sigma^2_{\gamma}-(t-2)\sigma^2_c \sigma^2_{\gamma}/f+\sigma^2_n}{\sigma^2_n \sigma^2_{\gamma}} \Ibb. \end{align} This implies that 
\begin{align}
H_t^{-1} &= \frac{\sigma^2_n \sigma^2_{\gamma}}{(t-1)\sigma^2_{\gamma}-(t-2)\sigma^2_c \sigma^2_{\gamma}/f+\sigma^2_n}\Ibb \nonumber \end{align} whereby  we get \begin{align}
    R_t &= \frac{f\Bigl((t-1)\sigma^2_{\gamma} +\sigma^2_n+\sigma^2_c \Bigr)}{\sigma^2_c((t-1)\sigma^2_{\gamma}\sigma^2_n+\sigma^2_c\sigma^2_{\gamma}+f\sigma^2_n)} \Ibb \nonumber \quad \mbox{and} \\
    R_t^{-1}&= \frac{\sigma^2_c((t-1)\sigma^2_{\gamma}\sigma^2_n+\sigma^2_c\sigma^2_{\gamma}+f\sigma^2_n)}{f\Bigl((t-1)\sigma^2_{\gamma} +\sigma^2_n+\sigma^2_c \Bigr)} \Ibb =b_t \Ibb \label{eq:Rtinv}. \end{align} Noting that 
    $M_t = \frac{f}{\sigma^2_n \sigma^2_c} \Ibb$ we then have \begin{align}
R_t^{-1}M_t&=\frac{(t-1)\sigma^2_{\gamma}\sigma^2_n+\sigma^2_c\sigma^2_{\gamma}+f\sigma^2_n}{\sigma^2_n\Bigl((t-1)\sigma^2_{\gamma} +\sigma^2_n+\sigma^2_c \Bigr)}\Ibb = \frac{(t-1)\sigma^2_{\gamma}\sigma^2_n+\sigma^2_c\sigma^2_{\gamma}+f\sigma^2_n}{(t-1)\sigma^2_{\gamma} \sigma^2_n+f\sigma^2_n } \Ibb =(1+ \frac{\sigma^2_c\sigma^2_{\gamma}}{(t-1)\sigma^2_{\gamma} \sigma^2_n+f\sigma^2_n })\Ibb.
\end{align}

Subsequently, we get that for $t>1$,
\begin{align}
\log\Bigl( {\rm{det}}(R_t^{-1}M_t) \Bigr)=d \log \Bigl( 1+ \frac{\sigma^2_c\sigma^2_{\gamma}}{(t-1)\sigma^2_{\gamma} \sigma^2_n+f\sigma^2_n }\Bigr) \leq d \log \Bigl( 1+ \frac{\sigma^2_c/ \sigma^2_n}{(t-1)} \Bigr)\leq \frac{d\sigma^2_c/ \sigma^2_n}{t-1},
\end{align}
whereby 
\begin{align}
   \sum_{t=1}^T \log\Bigl( {\rm{det}}(R_t^{-1}M_t) \Bigr) &\leq d \log \Bigl(1+ \frac{\sigma^2_c\sigma^2_{\gamma}}{f\sigma^2_n }\Bigr) + \sum_{t>1}^T \frac{d\sigma^2_c/ \sigma^2_n}{t-1} \\
   & \leq \frac{d\sigma^2_c}{\sigma^2_n} \Bigl(\frac{\sigma^2_{\gamma}}{f}+ \log (T-1)\Bigr)
\end{align}where the last inequality follows since $\log(1+x)\leq x$ and $\sum_{s=1}^T \frac{1}{s} \leq \log(T).$
\section{Details on  Experiments}\label{app:experiments}
In this section, we present details on the baselines implemented for stochastic CBs with unobserved true contexts. 
\subsection{Gaussian Bandits}
For Gaussian bandits, we implemented the baselines as explained below.
\paragraph{{\it TS\_naive}:} This algorithm implements the following action policy at each iteration $t$,
\begin{align*}
    a_t = \arg \max_{a \in \Acal} \phi(a,\chat_t)^\top \theta_t, 
\end{align*} where $\theta_t$ is sampled from a Gaussian distribution $\Nscr(\mu_{t-1,{\rm naive}},\Sigma_{t-1,{\rm naive}}^{-1}) $ with 
\begin{align*}
  \Sigma_{t-1,{\rm naive}} &= \frac{\Ibb}{\lambda}+\frac{1}{\sigma^2}\sum_{\tau=1}^{t-1}\phi(a_{\tau},\chat_{\tau})\phi(a_{\tau},\chat_{\tau}) ^\top \\
\mu_{t-1,{\rm naive}} &= \frac{\Sigma_{t-1,{\rm naive}}^{-1}}{\sigma^2} \Bigl(\sum_{\tau=1}^{t-1}r_{\tau}\phi(a_{\tau},\chat_{\tau})\Bigr). 
\end{align*}

\paragraph{{\it TS\_oracle}:} In this baseline, the agent has knowledge of the true predictive distribution $P(c_t|\chat_t,\gammastar)$. Consequently, at  each iteration $t$, the algorithm chooses action
\begin{align*}
    a_t = \arg \max_{a \in \Acal} \psi(a,\chat_t|\gammastar)^\top \theta_t, 
\end{align*} where $\theta_t$ is sampled from a Gaussian distribution $\Nscr(\mu_{t-1,{\rm poc}},\Sigma_{t-1,{\rm poc}}^{-1}) $ with 
\begin{align*}
  \Sigma_{t-1,{\rm poc}} &= \frac{\Ibb}{\lambda}+\frac{1}{\sigma^2}\sum_{\tau=1}^{t-1}\psi(a_{\tau},\chat_{\tau}|\gammastar)\psi(a_{\tau},\chat_{\tau}|\gammastar) ^\top \\
\mu_{t-1,{\rm poc}} &= \frac{\Sigma_{t-1,{\rm poc}}^{-1}}{\sigma^2} \Bigl(\sum_{\tau=1}^{t-1}r_{\tau}\psi(a_{\tau},\chat_{\tau}|\gammastar)\Bigr). 
\end{align*}
For Gaussian bandits, the following figure shows additional experiment comparing the performance of our proposed Algorithm~1 for varying values of the number $K$ of actions. All parameters are set as in Fig~\ref{fig:expt1}(Left).
\begin{figure}[h!]
    \centering
    \includegraphics[scale=0.4, clip=true, trim=0in 0in 0in 0.3in]{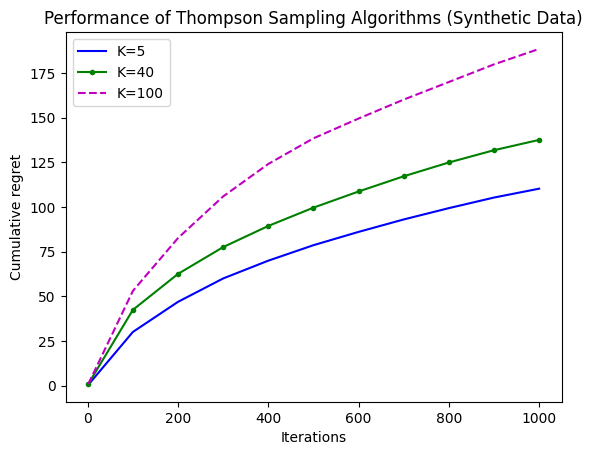}
    \caption{Bayesian cumulative regret of Algorithm 1 as a function of iterations over varying number $K$ of actions.}
    \label{fig:enter-label}
\end{figure}
\begin{comment}
\paragraph{{\it TS\_kk}:} This algorithm implements the following action policy at each iteration $t$,
\begin{align*}
    a_t = \arg \max_{a \in \Acal} \Ebb_{P(c_t)}[\phi(a,c_t)]^\top \theta_t, 
\end{align*} where $\theta_t$ is sampled from a Gaussian distribution $\Nscr(\mu_{t-1,{\rm kk}},\Sigma_{t-1,{\rm kk}}^{-1}) $ with 
\begin{align*}
  \Sigma_{t-1,{\rm kk}} &= \frac{\Ibb}{\lambda}+\frac{1}{\sigma^2}\sum_{\tau=1}^{t-1}\Ebb_{P(c_{\tau})}[\phi(a,c_{\tau})]\Bigl(\Ebb_{P(c_{\tau})}[\phi(a_{\tau},c_{\tau})] \Bigr)^\top \\
\mu_{t-1,{\rm kk}} &= \frac{\Sigma_{t-1,{\rm kk}}^{-1}}{\sigma^2} \Bigl(\sum_{\tau=1}^{t-1}r_{\tau}\Ebb_{P(c_{\tau})}[\phi(a,c_{\tau})]\Bigr). 
\end{align*}
\end{comment}
\subsection{Logistic Bandits}
In the case of logistic bandits, we implemented the baselines as explained below.
\paragraph{{\it TS\_naive}:} 
This algorithm considers the following sampling distribution:
\begin{align*}
    Q(\thetastar|\Hscr_{t-1,r,a,\chat}) \propto P(\thetastar) \prod_{\tau=1}^{t-1} {\rm Ber}(\mu(\phi(a_{\tau},\chat_{\tau})^{\top}\thetastar)).
\end{align*} However, due to the non-conjugateness of Gaussian prior $P(\thetastar)$ and Bernoulli reward likelihood, sampling from the above posterior distribution is not straightforward. Consequently, we adopt the Langevin Monte Carlo (LMC) sampling approach from \cite{xu2022langevin}. 
To sample $\theta_t$
at iteration $t$,  we run LMC for $I=50$ iterations with learning rate $\eta_t=0.2/t$ and inverse temperature parameter $\beta^{-1}=0.001$. Then, $\theta_t$ is chosen as the output of the LMC after $I=50$ iterations. Using the sampled $\theta_t$, the algorithm then chooses the action $a_t$ as
\begin{align*}
    a_t = \arg \max_{a \in \Acal} \phi(a,\chat_t)^\top \theta_t.
\end{align*}

\paragraph{{\it TS\_oracle}:} 
This algorithm considers the following sampling distribution:
\begin{align*}
    Q(\thetastar|\Hscr_{t-1,r,a,\chat}) \propto P(\thetastar) \prod_{\tau=1}^{t-1} {\rm Ber}(\mu(\psi(a_{\tau},\chat_{\tau}|\gammastar)^{\top}\thetastar)),
\end{align*} where $\psi(a_t,\chat_t|\gammastar)=\Ebb_{P(c_t|\chat_t,\gammastar)}[\phi(a_t,c_t)]$ is the expected feature map under the posterior predictive distribution with known $\gammastar$. As before, to sample from the above distribution, we use $I=50$ iterations of LMC with learning rate $\eta_t=0.2/t$ and inverse temperature parameter $\beta^{-1}=0.001$. Using the sampled $\theta_t$, the algorithm then chooses the action $a_t$ as
\begin{align*}
    a_t = \arg \max_{a \in \Acal} \psi(a,\chat_t|\gammastar)^\top \theta_t.
\end{align*}

\end{document}